\documentclass[final,12pt]{clear2026} 

\title[The Generalised Kernel Covariance Measure]{The Generalised Kernel Covariance Measure}
\usepackage{times}
\usepackage{bbm}
\usepackage{bm}
\usepackage{setspace}
\usepackage{enumitem}
\usepackage{booktabs}
\usepackage{caption}
\usepackage{mathrsfs}
\newcommand{\indep}{\perp\mkern-10mu\perp}

\newcommand{\E}{\mathbb{E}}
\newcommand{\R}{\mathbb{R}}

\newcommand{\X}{\mathcal{X}}
\newcommand{\Y}{\mathcal{Y}}
\newcommand{\Z}{\mathcal{Z}}
\newcommand{\B}{\mathcal{B}}

\newcommand{\F}{\mathcal{F}}
\newcommand{\G}{\mathcal{G}}
\newcommand{\Hc}{\mathcal{H}}

\newcommand{\tbnote}[1]{\textsuperscript{#1}}

\DeclareMathOperator*{\Cov}{Cov}

\DeclareMathOperator{\HS}{HS}

\DeclareMathOperator*{\argmin}{arg\,min}

\clearauthor{%
  \Name{Luca Bergen} \Email{bergen@leibniz-bips.de}\\
  \addr Leibniz Institute for Prevention Research and Epidemiology -- BIPS, Bremen, Germany \\
  Faculty of Mathematics and Computer Science, University of Bremen, Bremen, Germany
  \AND
  \Name{Dino Sejdinovic} \Email{dino.sejdinovic@adelaide.edu.au}\\
  \addr School of Mathematical Sciences and Australian Institute for Machine Learning \\
  Adelaide University, Adelaide, Australia
  \AND
  \Name{Vanessa Didelez} \Email{didelez@leibniz-bips.de}\\
  \addr Leibniz Institute for Prevention Research and Epidemiology -- BIPS, Bremen, Germany \\
  Faculty of Mathematics and Computer Science, University of Bremen, Bremen, Germany
}

\begin{document}

\maketitle

\begin{abstract}%
We consider the problem of conditional independence (CI) testing and adopt a kernel-based approach. Kernel-based CI tests embed variables in reproducing kernel Hilbert spaces, regress their embeddings on the conditioning variables, and test the resulting residuals for marginal independence. This approach yields tests that are sensitive to a broad range of conditional dependencies. Existing methods, however, rely heavily on kernel ridge regression, which is computationally expensive when properly tuned and yields poorly calibrated tests when left untuned, which limits their practical usefulness. We propose the Generalised Kernel Covariance Measure (GKCM), a regression-model-agnostic kernel-based CI test that accommodates a broad class of regression estimators. Building on the Generalised Hilbertian Covariance Measure framework \citep{lundborg_conditional_2022}, we characterise conditions under which GKCM satisfies uniform asymptotic level guarantees. In simulations, GKCM paired with tree-based regression models frequently outperforms state-of-the-art CI tests across a diverse range of data-generating processes, achieving better type I error control and competitive or superior power.
  
\end{abstract}

\begin{keywords}%
  Conditional independence testing, kernel conditional independence test, causal discovery%
\end{keywords}


 

\section{Introduction} \label{sec:bg}

Conditional independence (CI) is a central notion in probability theory and statistics \citep{dawid_conditional_1979}. It admits several equivalent characterisations, cf. \citet{constantinou_extended_2017} for details. We will use the following: for random variables $X,Y,Z$ with values in $(\X, \B_\X)$, $(\Y, \B_\Y)$, and $(\Z, \B_\Z)$, respectively, we say that $X$ is conditionally independent of $Y$ given $Z$, denoted by $X \indep Y \mid Z$, if
\begin{equation} \label{eq:CI}
    \operatorname{Cov}(f(X),g(Y) \mid Z) = \E[f(X)g(Y)\mid Z] -  \E[f(X)\mid Z] \E[g(Y)\mid Z] = 0
\end{equation}
almost surely for all bounded and measurable functions $f: \X \to \R$ and $g: \Y \to \R$. 
Conditional independence can be inferred from data via CI tests, i.e., statistical tests of the null hypothesis $X \indep Y \mid Z$. Such tests play a key role in causal inference, for example in constraint-based causal discovery \citep{glymour_review_2019} and invariant causal prediction \citep{peters_causal_2016, heinze-deml_invariant_2018}. 

Different methods are used for CI testing in practice, even though they may formally target a null hypothesis that is larger than the null of conditional independence. Many of these fall into two groups, which we refer to as residual- and kernel-based tests. In the following we briefly review these methods before introducing GKCM.

\subsection{Residual-based tests}

Throughout this section, let $X$ and $Y$ be square-integrable and let $\X \subseteq \R^p$ and $\Y \subseteq \R^q$. The first group of methods, which we call residual-based tests, is based on testing whether the mean conditional covariance matrix
\begin{equation} \label{eq:mccc}
    \E_Z[\Cov(X,Y \mid Z)] = \E[(X - \E[X \mid Z])(Y - \E[Y \mid Z])^\intercal] =0,
\end{equation}
which is implied by CI. Well-known examples are tests of vanishing partial covariance  under the assumption of joint normality, where the conditional expectations $\E[X \mid Z]$ and $\E[Y \mid Z]$ are estimated by least-squares linear regression models \citep[see, e.g.,][Section 15.5]{anderson_introduction_2003}. Another example is the Generalised Covariance Measure \citep[GCM,][]{shah_hardness_2020}, which avoids overly restrictive assumptions like joint normality. Instead, it allows the conditional expectations to have complex nonlinear dependencies on $Z$, which may be estimated using flexible regression or machine-learning methods. Unlike most CI tests, GCM can achieve uniform rather than merely pointwise asymptotic level over subsets of the null where the prediction errors of the regression models vanish sufficiently fast.\footnote{For the distinction between uniform and pointwise asymptotic level, see, e.g., \citet[][Section 1]{shah_hardness_2020}.} 

However, without strong parametric assumptions like joint normality, residual-based tests are limited by the fact that there exist distributions from the alternative satisfying $\E_Z[\Cov(X,Y \mid Z)] = 0$, which renders the conditional dependence 
undetectable to the tests. This limitation occurs because the condition \eqref{eq:mccc} is strictly weaker than CI as characterised by (\ref{eq:CI})  in two distinct respects. First, the tests do not consider the bounded functions of $X$ and $Y$, but instead (implicitly) the linear functions, since
\begin{equation*}
    \E_Z[\operatorname{Cov}(X,Y \mid Z)] = 0 \iff \E_Z[\operatorname{Cov}(\mathrm{u}^\intercal X, \mathrm{v}^\intercal Y \mid Z)] = 0
\end{equation*}
for all $\mathrm{u} \in \R^p$ and $\mathrm{v} \in \R^q$. Second, the tests do not assess whether the conditional covariances vanish almost surely, but only whether they vanish in expectation over $Z$ for all pairs of functions.  
Hence, even if the tests were to use sufficiently large function classes, they still could not detect conditional dependence under distributions where the mean conditional covariances vanish for all pairs of functions. We call methods targeting the mean conditional covariances mean-zero (as opposed to a.s.-zero) tests regardless of the function spaces considered. When considering all square-integrable functions of $X$ and $Y$, mean-zero testing corresponds to testing for weak CI \citep{daudin_partial_1980}.

Motivated by these shortfalls \cite{scheidegger_weighted_2022} proposed the weighted GCM (wGCM) and \cite{lundborg_projected_2024} the projected covariance measure (PCM). Both methods allow for the use of arbitrary regression models and uniform level guarantees (under the required assumptions), while being able to detect conditional dependence under larger subsets of the alternative. The wGCM tests whether $\Cov(X,Y \mid Z) = 0$ almost surely by weighting the regression residuals with a function of $Z$. For scalar $X$, PCM tests for conditional mean independence, i.e.,
\begin{equation*}
    \E[X\mid Y, Z] = \E[X\mid Z] \iff \E_Z[\operatorname{Cov}(X, h(Y,Z) \mid Z)] = 0
\end{equation*}
for all square-integrable functions $h$. It does so by replacing $Y$ in \eqref{eq:mccc} with a scalar-valued projection $f(Y,Z)$, which is regressed on $Z$. Testing for conditional mean independence is equivalent to a.s.-zero testing while additionally considering nonlinear functions of $Y$. 
Yet both tests only consider linear functions (at least) of $X$.

\subsection{Kernel-based tests}

The second group of methods, which we call kernel-based tests, combines residual-based testing with a large number of transformations by embedding the random variables in reproducing kernel Hilbert spaces as high-dimensional feature spaces. 
Let $\F \subseteq \R^{\X}$. A Hilbert space $(\F, \langle \cdot,  \cdot\rangle_\F)$ is called a reproducing kernel Hilbert space (RKHS) if there exists a symmetric positive definite function $k: \X \times \X \to \mathbb{R}$ satisfying $k(\cdot,x) \in \mathcal{F}$ for all $x \in \mathcal{X}$, and $\langle k(\cdot,x), f \rangle_{\mathcal{F}} = f(x)$ for all $x \in \mathcal{X}$ and $f \in \F$ (the \emph{reproducing property}). When $k$ satisfies these properties, it is called the reproducing kernel of $(\F, \langle \cdot, \cdot \rangle_\F)$ (see, e.g., \citealp{berlinet_reproducing_2004}). The reproducing kernel is used to map $\X \to \F$ via the canonical feature map $\phi: x \mapsto k(\cdot,x)$. 
Depending on the kernel and domain, each mapping or feature vector $\phi(x)$ may encode an infinite number of non-linear transformations of $x$. By the reproducing property 
the feature vectors satisfy
\begin{equation} \label{eq:trick}
    \langle \phi(x), \phi(x') \rangle_{\mathcal{F}}
    =k(x,x') \quad  \forall x, x' \in \mathcal{X},
\end{equation}
which enables the implicit evaluation of their inner products via the kernel. Algorithms are typically designed only to rely on these inner products, which allows them to avoid evaluating the high-dimensional feature vectors (commonly known as the \emph{kernel trick}). 

Examples of kernel-based CI tests include the seminal Kernel CI Test \citep[KCIT,][]{zhang_kernel-based_2011}, the Kernel Regression with Subsequent Independence Test \citep[KRESIT,][]{zhang_feature--feature_2017}, the Randomized Conditional Independence Test \citep[RCIT,][]{strobl_approximate_2019}, and the Randomized Correlation Test \citep[RCoT,][]{strobl_approximate_2019}. In all four methods, the variables are embedded into RKHSs and the mean conditional covariance of the embeddings is estimated via kernel ridge regression (see Sections \ref{sec:idea} and \ref{sec:KRR}). The tests differ mainly in which variables are embedded and in the dimensions of the RKHSs used: whereas KRESIT and RCoT embed $X$ and $Y$, KCIT and RCIT embed $(X,Z)$ and $Y$, similar to PCM. Accordingly, the former are mean-zero and the latter are a.s.-zero tests. Furthermore, KCIT and KRESIT use Gaussian kernels which induce infinite-dimensional RKHSs, whereas RCIT and RCoT approximate these kernels using Random Fourier Features \citep{rahimi_random_2007}, which induce finite-dimensional RKHSs. Therefore RCIT and RCoT may be seen as faster, approximate versions of KCIT and KRESIT, respectively.

Using the kernel trick, kernel-based tests are able to consider a large number of transformations of $X$ and $Y$ simultaneously. This enables the tests to detect conditional dependencies under large subsets of the alternative. By using suitable kernels they can also accommodate different variable types and mixed data. 
However, even though there exists a wide range of RKHS-valued regression methods, 
existing kernel-based CI tests exclusively use kernel ridge regression. This is detrimental to their performance, since kernel ridge regression is sensitive to the choice of hyperparameters and tuning the hyperparameters is computationally costly. Thereby the tests become either computationally prohibitive or unreliable when hyperparameter tuning is bypassed. 
Furthermore, to the best of our knowledge, it has not been investigated whether asymptotic uniform type-I error guarantees can be shown to hold for kernel-based tests. 
This is subpar compared to the residual-based tests, where sufficient conditions have been stated without prior restriction of the regression methods to be used in testing \citep{shah_hardness_2020,scheidegger_weighted_2022,lundborg_projected_2024}. 

\subsection{The key idea of GKCM} \label{sec:idea}

To resolve these issues, we propose the Generalised Kernel Covariance Measure (GKCM). Our method is similar to the existing kernel-based CI tests in procedure, especially to KRESIT, but allows for the use of arbitrary Hilbert space-valued regression methods. Furthermore, we define GKCM as a Generalised Hilbertian Covariance Measure \citep[GHCM,][]{lundborg_conditional_2022}, which allows us to state conditions for uniform asymptotic level guarantees. GHCM tests are a class of CI tests for Hilbert space-valued random variables introduced by \citet{lundborg_conditional_2022}. The authors show that GHCM tests have uniform asymptotic type-I error control under assumptions similar to GCM; most importantly, the in-sample prediction errors of the regression methods need to vanish sufficiently fast uniformly over the subset of distributions. While the GHCM framework has originally been developed for CI testing in functional data, we use it instead on the RKHS embeddings to infer CI between $X$ and $Y$. We show that, under mild assumptions and for many choices of kernels, $X$ and $Y$ are conditionally independent if and only if their embeddings are conditionally independent, which justifies our approach.

For a first idea of GKCM, let $(\F, \langle \cdot, \cdot\rangle_\F)$ and $(\G, \langle \cdot, \cdot\rangle_\G)$ denote RKHSs with reproducing kernels $k: \X \times \X \to \R$ and $l: \Y \times \Y \to \R$, and canonical feature maps defined by $\phi(x) := k(\cdot, x)$ and $\varphi(y) := l(\cdot, y)$, respectively. 
GKCM targets the mean conditional covariance of $\phi(X)$ and $\varphi(Y)$ given $Z$ defined as the operator
\begin{equation*} \label{eq:MCCoperator}
    \mathbf{C}_{XY \boldsymbol{\cdot} Z} := \E\big[\big(\phi(X) - \E[\phi(X) \mid Z]\big)  \otimes \big(\varphi(Y) - \E[\varphi(Y) \mid Z]\big)\big], 
\end{equation*}
where $\phi(X)$ and $\varphi(Y)$ as well as their conditional expectations 
are random variables taking values in $\F$ and $\G$, respectively, and for any $f \in \F$ and $g \in \G$, the rank-one operator $f\otimes g: \G \to \F$ is defined by $(f \otimes g)(g') := \langle g, g'\rangle_\G f$ analogously to the outer product $(\mathrm{u}\mathrm{v}^\intercal)\mathrm{v}' = (\mathrm{v} ^\intercal \mathrm{v}')\mathrm{u}$ in Euclidean space \citep{pogodin_practical_2025, park_measure-theoretic_2020}. 
Since GKCM tests whether
\begin{equation*} \label{eq:weakCI}
    \mathbf{C}_{XY \boldsymbol{\cdot} Z} = 0 \iff \E_Z[\Cov(f(X), g(Y) \mid Z)] = 0 \quad \forall f \in \F, g \in \G,
\end{equation*} 
we expect it to detect conditional dependencies under large subsets of the alternative when using kernels with sufficiently rich RKHSs. In particular, if $k$ and $l$ are $L^2$-universal, i.e. $\F$ and $\G$ are dense in $L^2(\X, \mu)$ and $L^2(\Y, \nu)$ with respect to the $L^2$-norm for all Borel probability measures $\mu$ and $\nu$ \citep{sriperumbudur_universality_2011}, the condition $\mathbf{C}_{XY \boldsymbol{\cdot} Z} = 0$ is equivalent to weak CI. Table \ref{tab:CIcomp} compares GKCM to the other tests.

\begin{table}[ht]
    \small
    \centering
    \addtolength{\tabcolsep}{-3.6pt}
    \begin{tabular}{lcccccccc}
        \toprule &
        \textbf{GCM} & \textbf{wGCM} & \textbf{PCM} & \textbf{KCIT} & \textbf{KRESIT} & \textbf{RCIT} & \textbf{RCoT} & \textbf{GKCM} \\
        \midrule
        \textbf{Regression methods} & Any & Any & Any & KRR & KRR & KRR\tbnote{$\dagger$} & KRR\tbnote{$\dagger$} & Any \\
        \textbf{Type-I error control} & Uniform & Uniform & Uniform & Pointw. & Pointw. & Pointw. & Pointw. & Uniform \\
        \textbf{Function spaces} & Linear & Linear & Mixed & RKHS & RKHS & RKHS\tbnote{$\dagger$} & RKHS\tbnote{$\dagger$} & RKHS \\
        \textbf{Conditional covariances} & Mean-z. & A.s.-zero & A.s.-zero & A.s.-zero & Mean-z. & A.s.-zero & Mean-z. & Mean-z. \\
        
        \bottomrule
        \multicolumn{9}{l}{\vspace{5pt}\footnotesize
        \tbnote{$\dagger$} Approximated using Random Fourier Features
        }
    \end{tabular}    
    \caption{Comparison of residual- and kernel-based CI tests}
    \label{tab:CIcomp}
\end{table}

Like GCM, GKCM is a mean-zero test. In contrast, KCIT uses a joint embedding of $(X,Z)$ to test for a.s.-zero conditional covariances. Under $L^2$-universal kernels, this is equivalent to testing the stronger null hypothesis of CI \citep{pogodin_practical_2025, he_hardness_2025}. Although the use of a joint embedding does not strengthen central assumptions required for asymptotic validity, it may inflate type-I error rates in finite samples (see Appendix \ref{sec:joint} for details and the discussion in \citealp{he_hardness_2025}). To better control finite-sample type-I error, we therefore refrain from using a joint embedding.

\section{Background and assumptions} \label{sec:bg}

Let $X,Y,Z$ be random variables on the measurable space $(\Omega, \mathcal{A})$ with values in measurable spaces $(\X, \B_\X)$, $(\Y, \B_\Y)$, and $(\Z, \B_\Z)$, respectively, where $\X,\Y,\Z$ are topological spaces and $\B_\X, \B_\Y, \B_\Z$ denote their Borel $\sigma$-algebras. 
The measurable space $(\Omega, \mathcal{A})$ is equipped with a family of probability measures $(\mathbb{P}_P)_{P \in \mathcal{P}}$ such that the joint distribution of $(X,Y,Z)$ under $\mathbb{P}_P$ is $P$. 
We denote the null hypothesis by $\mathcal{P}_0 := \{P \in \mathcal{P}: X \indep Y \mid Z\}$. 
For $\tilde{\mathcal{P}} \subseteq \mathcal{P}$ and a sequence of real-valued 
random variables $(V_n)_{n \in \mathbb{N}}$, we write $V_n = o_{\tilde{\mathcal{P}}}(1)$ to denote that for all $\epsilon > 0$,
\begin{equation*}
    \lim_{n \to \infty}\sup_{P \in \tilde{\mathcal{P}}} \mathbb{P}_P
    (\lvert V_n \rvert \geq \epsilon ) = 0.
\end{equation*}
Furthermore, $\operatorname{HS}(\G,\F)$ denotes the set of Hilbert-Schmidt (HS) operators $\G \to \F$ \citep[Ch. 4.4]{hsing_theoretical_2015}, which has a Hilbert space structure when equipped with the inner product 
$\langle \mathbf{A}, \mathbf{B}\rangle_{\operatorname{HS}} := \sum_{j \in J} \langle \mathbf{A} g_j, \mathbf{B} g_j\rangle_\F$, where $(g_j)_{j \in J}$ is an arbitrary orthonormal basis of $\G$. 
In particular, for any $f \in \F$ and $g \in \G$,  we have $f \otimes g \in \HS(\G,\F)$. 

\paragraph{Assumptions} In the following, we assume that
 \begin{enumerate}[label=A.\arabic*,ref=A.\arabic*, itemsep = 0em]
     \item $\X, \Y, \Z$ are Polish spaces, \label{asm1}
     \item $k$ and $l$ are continuous, \label{asm2}
     \item $k$ and $l$ are bounded, i.e., $\kappa_X := \sup_{x \in \X} k(x,x) < \infty$ and $\kappa_Y := \sup_{y \in \Y} l(y,y) < \infty$, \label{asm3}
     \item $\phi$ and $\varphi$ are injective. \label{asm4}
 \end{enumerate}
These assumptions ensure that all population quantities from the previous section are well defined, that the regularity assumptions of the GHCM framework \citep[Section 1.2]{lundborg_conditional_2022} are satisfied, and that testing for $\phi(X) \indep \varphi(Y) \mid Z$ is equivalent to testing for $X \indep Y \mid Z$ since the two CI statements coincide, as we argue below.

Since $\X$ and $\Y$ are separable by assumption \ref{asm1}, it follows from \ref{asm2} and \citet[Lemma 4.33]{steinwart_support_2008} that $\F$ and $\G$ are separable. As $(\F, \langle \cdot, \cdot \rangle_\F)$ and $(\G, \langle \cdot, \cdot \rangle_\G)$ are complete, the separability of $\F$ and $\G$ implies that they are Polish when endowed with their norm topologies. Moreover, by \ref{asm2} and \citet[Lemma 4.29]{steinwart_support_2008}, $\phi$ and $\varphi$ are continuous and therefore Borel measurable. 
Hence $\phi(X)$ and $\varphi(Y)$ are random variables on $(\Omega, \mathcal{A})$ taking values in the standard Borel spaces $(\F, \B_\F)$ and $(\G, \B_\G)$, respectively. 

Since $\phi(X)$ and $\varphi(Y)$ are Borel measurable and $\F$ and $\G$ are separable, $\phi(X)$ and $\varphi(Y)$ are strongly measurable. 
By assumption \ref{asm3} $\E[\lVert \phi(X)\rVert^2_\F]= \E[k(X,X)]  < \infty$ and $\E[\lVert \varphi(Y)\rVert^2_\G] = \E[l(Y,Y)]< \infty$ for all $P \in \mathcal{P}$. Hence $\phi(X)$ and $\varphi(Y)$ are square-integrable and therefore Bochner integrable \citep[Appendix E]{cohn_measure_2013}. Therefore the conditional expectations $\E[\phi(X) \mid Z]$ and $\E[\varphi(Y) \mid Z]$ (henceforth conditional mean embeddings) exist as $\sigma(Z)$-measurable random variables \citep{park_measure-theoretic_2020}. 
By the Doob-Dynkin Lemma \citep[Lemma 1.14]{kallenberg_foundations_2021}, there exist Borel-measurable functions $F_P: \Z \to \F$ and $G_P: \Z \to \G$ satisfying $\E[\phi(X) \mid Z] = F_P(Z)$ and $\E[\varphi(Y) \mid Z] = G_P (Z)$ almost surely for each $P \in \mathcal{P}$. 

Assumptions \ref{asm1}, \ref{asm2}, and \ref{asm4} ensure that CI is preserved under $\phi$ and $\varphi$. By \citet[Proposition 2.3]{constantinou_extended_2017} $X \indep Y \mid Z$ is equivalent to $\E[\mathbbm{1}_{A \cap B} \mid Z] = \E[\mathbbm{1}_{A} \mid Z] \E[\mathbbm{1}_{B} \mid Z]$ almost surely for all $A \in \sigma(X)$ and $B \in \sigma(Y)$. Hence 
\begin{equation*}
    X \indep Y \mid Z \iff \phi(X) \indep \varphi(Y) \mid Z
\end{equation*}
if $\sigma(\phi(X))=\sigma(X)$ and $\sigma(\varphi(Y))=\sigma(Y)$, which follows from Lemma \ref{lem:sigequal} (proved in Appendix \ref{sec:proof}). 
\begin{lemma} \label{lem:sigequal}
Let $\X$ and $\F$ denote Polish spaces. For every random variable $X$ taking values in the standard Borel space $(\X, \B_X)$ and every Borel-measurable and injective function $\phi: \X \to \F$, it holds that $\sigma(X) = \sigma(\phi(X))$.
\end{lemma}
All of the assumptions are mild and can be shown to hold in many settings. Assumptions \ref{asm1} -- \ref{asm4} hold, e.g., for $\R^d$ and open or closed subsets equipped with Gaussian kernels, 
or for finite sets equipped with the Dirac kernel. 

\section{GKCM: Definition and Properties}

To formally define GKCM, let $\{(X_i, Y_i, Z_i)\}_{i=1}^n$ be an i.i.d.\ sample of $(X,Y,Z)$, and let $\hat{F}_n: \Z \to \F$, $\hat{G}_n: \Z \to \G$ denote regression functions trained in-sample. Define the centred residuals
\begin{equation*}
    \hat{\varepsilon}_i := \phi(X_i) - \hat{F}_n(Z_i) - \hat{\mu}_\varepsilon, \quad
    \hat{\xi}_i := \varphi(Y_i) - \hat{G}_n(Z_i) - \hat{\mu}_\xi,
\end{equation*}
\begin{equation*}
    \hat{\mu}_\varepsilon := \frac{1}{n}\sum_{i=1}^n (\phi(X_i) - \hat{F}_n(Z_i)), \quad
    \hat{\mu}_\xi := \frac{1}{n} \sum_{i=1}^n (\varphi(Y_i) - \hat{G}_n(Z_i)).
\end{equation*}
The resulting empirical mean conditional covariance operator is
\begin{equation*}
    \hat{\mathbf{C}}_{XY \cdot Z}^{(n)} := \frac{1}{n} \sum_{i=1}^n \hat{\varepsilon}_i  \otimes \hat{\xi}_i,
\end{equation*}
and the test statistic of GKCM is defined as
\begin{equation*}
    T_n := n \lVert \hat{\mathbf{C}}_{XY \cdot Z}^{(n)}\rVert^2_{\text{HS}}. 
\end{equation*}
Note that analogous test statistics are used in KCIT and KRESIT as well as in the GHCM framework, which allows us to use properties established for the latter in the following.

\citet{lundborg_conditional_2022} show that over subsets of the null hypothesis where the in-sample prediction errors of the regression models vanish sufficiently fast and additional moment conditions are met, the scaled operator $\sqrt{n}\hat{\mathbf{C}}^{(n)}_{XY \cdot Z}$ converges uniformly in distribution to a mean-zero Gaussian random element \citep[Section 2.3]{da_prato_stochastic_2014} with unknown covariance operator $\mathbf{C}_P$. 
As a consequence, the asymptotic null distribution of the quadratic form $T_n = \lVert \sqrt{n}\hat{\mathbf{C}}^{(n)}_{XY \cdot Z} \rVert_{\HS}^2$ is characterised by the eigenvalues of $\mathbf{C}_P$. These are consistently estimated by the non-zero eigenvalues $(\lambda_i)_{i = 1}^d$ of the matrix
\begin{equation*}
    \mathrm{T} := \frac{1}{(n-1)}\mathrm{H}\mathrm{R}\mathrm{H},
\end{equation*}
where $\mathrm{R}\in \R^{n \times n}$ with $ \mathrm{R}_{ij} := \langle \hat{\varepsilon}_i, \hat{\varepsilon}_j\rangle_\F \langle \hat{\xi}_i, \hat{\xi}_j \rangle_\G$, and
$\mathrm{H}:= \mathrm{I}_n - n^{-1}\mathrm{J}_n$. The null distribution of $T_n$ is then approximated by the generalised chi-square distribution of $\sum_{i=1}^d \lambda_i V_i^2$, where $V_i$ are i.i.d.\ standard normal random variables, and the level-$\alpha$ test function is defined as
\begin{equation*}
    \tau_n := \mathbbm{1}\{T_n \geq q_\alpha \},
\end{equation*}
where $q_\alpha$ denotes the $1-\alpha$ quantile of $\sum_{i=1}^d \lambda_i V_i^2$. 
In practice the test statistic can be computed as $T_n = n^{-1} \sum_{i,j=1}^n \mathrm{R}_{ij}$ and the null distribution can be approximated using miscellaneous methods \citep[see, e.g.,][]{bodenham_comparison_2016}. 

\citet[Theorems 2 and 3]{lundborg_conditional_2022} state conditions for uniform convergence and for the resulting uniform level guarantees of GHCM tests. For completeness, these are reproduced below in Theorem \ref{thm:unif}. In order to state the condition and results we define the population residuals
\begin{equation*}
    \varepsilon_P := \phi(X) - \E[\phi(X) \mid Z], \quad \xi_P := \varphi(Y) - \E[\varphi(Y) \mid Z],
\end{equation*}
the conditional variances
    \begin{equation*}
    u_P(z) := \E[\lVert\varepsilon_P\rVert_\F^2\mid Z = z], \quad
    v_P(z) := \E[\lVert\xi_P\rVert_\G^2\mid Z = z],
    \end{equation*}
and the (weighted) in-sample mean square prediction errors 
\begin{equation*}
    \mathcal{E}_{n}^F := \frac{1}{n} \sum_{i=1}^n \lVert F_P(Z_i) - \hat{F}_n(Z_i) \rVert^2_\F, \quad
    \mathcal{E}_{n}^G := \frac{1}{n} \sum_{i=1}^n \lVert G_P(Z_i) - \hat{G}_n(Z_i) \rVert^2_\G,
\end{equation*}
\begin{equation*}
    \tilde{\mathcal{E}}_{n}^F := \frac{1}{n} \sum_{i=1}^n \lVert F_P(Z_i) - \hat{F}_n(Z_i) \rVert^2_\F v_P(Z_i), \quad
    \tilde{\mathcal{E}}_n^G := \frac{1}{n} \sum_{i=1}^n \lVert G_P(Z_i) - \hat{G}_n(Z_i) \rVert^2_\G u_P(Z_i). 
\end{equation*}
With the above arguments, we have the following result. 
\begin{theorem} \label{thm:unif}
    Let $\tilde{\mathcal{P}}_0 \subset \mathcal{P}_0$ such that
    \begin{enumerate}[label=B.\arabic*,ref=B.\arabic*, itemsep = 0em]
        \item $n \mathcal{E}_{n}^F \mathcal{E}_{n}^G = o_{\tilde{\mathcal{P}}_0}(1)$, \label{asm:rates}
        \item  $\tilde{\mathcal{E}}_{n}^F = o_{\tilde{\mathcal{P}}_0}(1)$ and $\tilde{\mathcal{E}}_{n}^G = o_{\tilde{\mathcal{P}}_0}(1)$, \label{asm:cons}
        \item $\inf_{P \in \tilde{\mathcal{P}}_0} \E[\lVert \varepsilon_P \rVert_\F^2 \lVert \xi_P \rVert_\G^2] > 0$ and $\sup_{P \in \tilde{\mathcal{P}}_0} \E[\lVert \varepsilon_P \rVert_{\F}^{2+\eta} \lVert \xi_P \rVert_{\G}^{2+\eta}] < \infty$ for some $\eta > 0$, 
        \item for some orthonormal bases $(f_i)_{i\in I}$ and $(g_j)_{j \in J}$ of $\F$ and $\G$, it holds that
        \begin{equation*}
            \lim_{K \to \infty}\sup_{P \in \tilde{\mathcal{P}}_0} \sum_{(i,j): i+j \geq K} \E[\langle f_i, \varepsilon_P\rangle_\F^2  \langle g_j, \xi_P\rangle_\G^2] = 0,
        \end{equation*}
        \item $\inf_{P \in \tilde{\mathcal{P}}_0} \lVert \mathbf{C}_P \rVert_{\operatorname{op}} > 0$.
    \end{enumerate}
    Then, for each $\alpha \in (0,1)$, the level-$\alpha$ GKCM test $\tau_n$ satisfies 
    \begin{equation*}
        \lim_{n \to \infty}\sup_{P \in \tilde{\mathcal{P}}_0} \lvert \mathbb{P}_P(\tau_n = 1) - \alpha \vert  = 0.
    \end{equation*}
\end{theorem}
Theorem \ref{thm:unif} implies that the central challenge in kernel-based CI testing consists in estimating the conditional mean embeddings. This conclusion has recently been stated independently for kernel-based CI tests based on U-statistics by \citet{he_hardness_2025}. Assumption \ref{asm:rates} specifies the (product) error rates required to hold for the regression models uniformly over $\tilde{\mathcal{P}}_0$. The assumption is satisfied, e.g., if $\sqrt{n}\mathcal{E}_{n}^F = o_{\tilde{\mathcal{P}}_0}(1)$ and $\sqrt{n}\mathcal{E}_{n}^G = o_{\tilde{\mathcal{P}}_0}(1)$, yet one model may converge slower if the other is able to compensate. Since the mean square prediction errors are defined in-sample, the regression models are not required to extrapolate well out-of-sample. By the discussion in \citet[Section 4.2]{lundborg_conditional_2022}, assumption \ref{asm:cons} is satisfied if $\mathcal{E}_{n}^F = o_{\tilde{\mathcal{P}}_0}(1)$ and $\mathcal{E}_{n}^G = o_{\tilde{\mathcal{P}}_0}(1)$, since $u_P(Z_i) \leq 4 \kappa_X$ and $v_P(Z_i) \leq 4 \kappa_Y$ almost surely.

\section{RKHS-valued regression}

Since the central challenge in kernel-based CI testing is to specify suitable regression models, we review the main approaches to this RKHS-valued or output-kernel regression problem. In particular, we focus on aspects most relevant in practice, namely modelling assumptions, robustness to model misspecification, tuning requirements, and computational cost.

\subsection{Kernel ridge regression} \label{sec:KRR}

A prominent subgroup consists of identity-decomposable input–output kernel regression (IOKR) methods, which are parametrised by an input kernel $m: \Z \times \Z \to \R$ \citep{brouard_input_2016}. In these methods, the feature map $\psi(z):= m(\cdot,z) \in \Hc$ is used as a high-dimensional feature expansion of the covariates. The regression coefficients are HS operators $\hat{\mathbf{C}}:\Hc \to \F$  learned via regularised empirical risk minimisation, yielding regression models of the form $\hat{F}_n(z) = \hat{\mathbf{C}} \psi(z)$. 
Examples include kernel ridge regression \citep{grunewalder_conditional_2012, li_optimal_2022}, robust regression methods \citep{laforgue_duality_2020}, 
and kernel principal component regression \citep{meunier_optimal_2024}. 

The most common IOKR method and the standard method in kernel-based CI testing is kernel ridge regression (KRR). 
For a fixed input kernel $m$ and regularisation parameter $\lambda > 0$, the KRR model is defined by $\hat{F}_n(z) = \hat{\mathbf{C}}_\lambda \psi(z)$ with coefficients
\begin{equation*} 
 \hat{\mathbf{C}}_\lambda := \argmin_{\mathbf{C} \in \HS(\Hc,\F) } \frac{1}{n} \sum_{i=1}^n \lVert \phi(x_i) - \mathbf{C} \psi(z_i)\rVert_\F^2 + \lambda \lVert \mathbf{C} \rVert_{\HS}^2.
\end{equation*}
The model admits the closed-form expression
\begin{equation} \label{eq:KRR}
    \hat{F}_n(z) = \frac{1}{n} \sum_{i=1}^n w_i(z) \phi(x_i), \quad \mathrm{w}(z) := (\mathrm{M}+ n \lambda \mathrm{I}_n)^{-1} \mathrm{m}(z) \in \R^n,
\end{equation}
where $\mathrm{M} \in \R^{n \times n}$ is a Gram matrix with $\mathrm{M}_{ij}:= m(z_i,z_j)$ and $\mathrm{m}(z) := [m(z_1,z), \dots, m(z_n,z)]^\intercal$. 

The performance of KRR is codetermined by the choice of input kernel. For covariates taking values in $\Z \subseteq \R^d$, the established kernel-based CI tests often use Gaussian tensor product kernels \citep{szabo_characteristic_2018} 
\begin{equation*}
    m(z,z') = \prod_{j=1}^d m_j(z_j, z_j'), \quad m_j(z_j, z_j') = \exp\left(- \frac{(z_j - z_j')^2}{2 \sigma_j^2} \right),
\end{equation*}
which are parametrised by lengthscales $\sigma_1, \dots, \sigma_d > 0$ (other parametric kernel families may be considered in addition).
The lengthscales and $\lambda$ are treated as hyperparameters and can be tuned jointly, e.g.\ using surrogate likelihood maximisation \citep{zhang_kernel-based_2011} or leave-one-out cross-validation \citep{pogodin_practical_2025}. 
While $\lambda$ controls the norm of the coefficients, the lengthscales control the effective size of the RKHS, since the RKHSs associated with Gaussian tensor-product kernels are nested in the lengthscales \citep[Propositions 3.5 and 5.2]{zhang_inclusion_2011}.

The input kernel matters because the population learning rate of KRR depends on the eigenvalue decay of the kernel integral operator, on an embedding property of the associated RKHS, and on the smoothness of the true regression function $F_P$ relative to the hypothesis space \citep{li_towards_2024}. Heuristically, enlarging the hypothesis space may improve approximation, whereas shrinking it may improve estimation. Whether a more restrictive hypothesis space is beneficial depends on whether it still contains sufficiently good approximations to the true regression function. In practice, however, this is typically unknown and cannot be assessed directly. 

At the same time, data-driven model selection is computationally demanding, since fitting a candidate model has computational cost growing cubically in $n$. Even for a fixed kernel family, choosing a separate lengthscale for each covariate plus $\lambda$ requires tuning $d+1$ hyperparameters. In many settings, this makes exhaustive tuning infeasible, so implementations often rely on heuristic choices instead. Since KRR is typically sensitive to the choice of tuning parameters, the covariate feature map often becomes a bottleneck in kernel-based CI testing, either computationally through tuning costs or statistically through suboptimal regression performance.

\subsection{Random Forests} \label{sec:RF}

The above practical issues motivate the use of alternative regression methods that do not rely on input kernels. Examples include RKHS-valued extensions of random forests \citep{geurts_kernelizing_2006, cevid_distributional_2022}, neural networks \citep{shimizu_neural-kernel_2024, el_ahmad_deep_2024} or gradient-boosted learners \citep{geurts_gradient_2007}. 

In the following, we will focus on random forests. Compared to KRR, they avoid the need to specify an input kernel and typically require much less hyperparameter tuning. This is well known in the scalar-valued setting and appears to hold for the RKHS-valued case as well. In our experiments, random forests often achieved strong performance without any parameter tuning (see Section \ref{sec:sim}). Finally, their representation of the covariates is more transparent: while KRR represents the covariates through a kernel feature map, regression trees represent them through split-induced regions of $\mathcal{Z}$.

In analogy to scalar-valued random forests, RKHS-valued random forests are bagged ensembles of binary RKHS-valued regression trees. Each tree is parametrised by a set of node-wise split rules of the covariates, 
which recursively partition $\Z$ into disjoint regions. For any $\mathcal{I}\subseteq \{1, ..., n\}$, denote $\Phi_\mathcal{I} := \{\phi(x_i)\}_{i \in \mathcal{I}}$.
The splits are chosen to maximise the reduction in the variance of the embeddings,
    \begin{equation*}
            \operatorname{Var}(\Phi_\mathcal{S}) - \frac{\lvert \mathcal{S}_l \rvert}{\rvert \mathcal{S} \rvert }\operatorname{Var}(\Phi_{\mathcal{S}_l}) - \frac{\lvert \mathcal{S}_r \rvert}{\rvert \mathcal{S} \rvert }\operatorname{Var}(\Phi_{\mathcal{S}_r}),
    \end{equation*}
where $\mathcal{S}$ is the index set of the subsample at the current split node and $\mathcal{S}_l$ and $\mathcal{S}_r$ are the corresponding left and right child nodes. In Output Kernel Random Forests \citep{geurts_kernelizing_2006} the variance is defined as
     \begin{align} \label{eq:var}
          \operatorname{Var}(\Phi_\mathcal{S}) := \frac{1}{\rvert \mathcal{S} \rvert} \sum_{i \in \mathcal{S}} \left\lVert \phi(x_i) - \hat{\mu}_{\mathcal{S}} \right\rVert^2_\F 
         = \frac{1}{\rvert \mathcal{S} \rvert} \sum_{i \in \mathcal{S}}  k(x_i, x_i)  - \frac{1}{\rvert \mathcal{S} \rvert^2} \sum_{i \in \mathcal{S} } \sum_{j \in \mathcal{S}} k(x_i,x_j),
     \end{align}
where $\hat{\mu}_{\mathcal{S}} := \ \rvert \mathcal{S} \rvert^{-1} \sum_{i \in \mathcal{S}} \phi(x_i)$. This criterion is evaluated by the kernel trick, while Distributional Random Forests \citep{cevid_distributional_2022} use a related criterion, in which the feature vectors in \eqref{eq:var} are approximated by Random Fourier Features to accelerate learning by avoiding the double sum in the kernel expression. Given $m$ trained regression trees, the random forest model can be represented by 
     \begin{equation*}
         \hat{F}_n(z) =\sum_{i=1}^n \tilde{w}_i(z) \phi(x_i), \quad \tilde{w}_i(z): =  \frac{1}{m} \sum_{j=1}^m\frac{ \mathbbm{1}\{z_i \in \mathcal{L}_j(z)\}}{\lvert \mathcal{L}_j(z) \rvert} \in [0,1],
     \end{equation*}
where $\mathcal{L}_j(z) \subseteq \{z_1, \dots, z_n\}$ denotes the subset of observations falling into the same terminal node or leaf as $z$ in the $j$th tree. 

\section{Simulation study} \label{sec:sim}

To illustrate the finite-sample performance of GKCM regarding level and power, we conduct a simulation study where we compare our method to GCM, wGCM and PCM, and to KCIT, RCIT and RCoT. The code is publicly available at GitHub.\footnote{https://github.com/lucabergen/GKCM} Since KRESIT is analogous to GKCM when using KRR (modulo computation of the p-value), we expect their performance to be very similar. Therefore we exclude KRESIT from the comparison. 

\paragraph{Methods and parameters}

To investigate the impact of different regression methods in kernel-based testing, we include GKCM using RKHS-valued random forests (i.e., Distributional Random Forests from the \texttt{R}-package \texttt{drf};  \citealp{michel_drf_2021}) and using KRR. We refer to them by GKCM RF and GKCM KRR, respectively. Denoting the number of conditioning variables by $p$, we set \texttt{num.trees} $= p \times 100$,  \texttt{mtry} $ = p$, and \texttt{min.node.size} $= 5$ to ensure that the random forests are sufficiently flexible. The hyperparameter values for KRR are described below.

For GCM, wGCM and PCM, we use the \texttt{R}-package \texttt{comets} \citep{kook_comets_2025} with random forests auto-tuned 
on \texttt{max.depth} $\times$ \texttt{mtry} with 500 trees for all regressions. This corresponds to using general-purpose, flexible regression models (as opposed to domain-informed regression models), which we consider to be a realistic use case for the residual-based tests in practice. 

RCIT, RCoT and KCIT are self-implemented for comparability. For all kernel ridge regressions (including GKCM KRR) we use a Gaussian input kernel on $Z$, 
set the lengthscale via the median heuristic \citep{garreau_large_2018}, and set $\lambda = 10^{-3}/n$. For KCIT, we use a U-statistic and the wild bootstrap as described in \citet[][Appendix H.1]{he_hardness_2025}, yet without using sample-splitting.

\paragraph{Data-generating processes}

\begin{figure}[h]
    \centering
    \includegraphics[width=\linewidth]{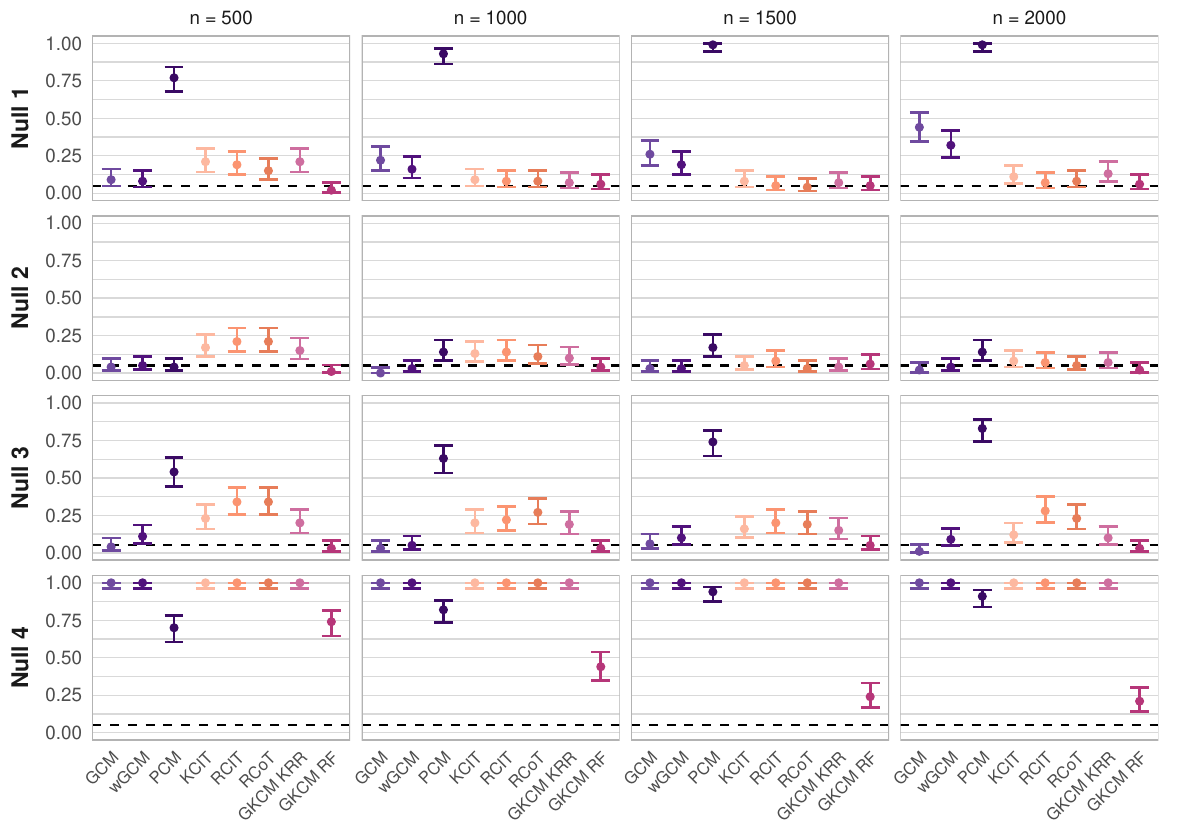}
    \caption{Rejection rates in the null settings with rejection threshold $p < 0.05$ (100 iterations). Error bars indicate 95\% Wilson confidence intervals and dashed lines the nominal level.}
\label{fig:null}
\end{figure}

We consider seven scenarios, four under the null and three under the alternative, with continuous $X,Y,Z$ and varying sample sizes ($n \in \{500, 1000, 1500, 2000\}$). In each setting we sample i.i.d.\ errors $(\varepsilon_X, \varepsilon_Y) \sim \mathcal{N}_2(0,\mathrm{I}_2)$ and conditioning variables $Z = (Z_1, \dots, Z_7)\sim \mathcal{N}_7(0,\mathrm{I}_7)$. The null settings include linear main effects, conditional means and variances with complex functional dependencies, post-nonlinear models \citep{zhang_identifiability_2009}, and a strong correlation between $X$ and $Y$. The results are shown in Figure \ref{fig:null}.
\begin{enumerate}[label=\textbf{Null \arabic*}:,ref=\arabic*, align=left, labelwidth=!, leftmargin=*, itemsep = 0em]
    \item \label{eq:null1}$
    \begin{aligned}[t]
    X &= 0.4Z_1 + 0.5Z_2 + 0.6 Z_3 - 0.7 Z_4 + Z_7 + \varepsilon_X, \\
    Y &= 0.6Z_1 - 0.2Z_2 + 0.3Z_4 + 0.9Z_5 - 0.5Z_6  + \varepsilon_Y
    \end{aligned}
    $
    \item \label{eq:null2}$
    \begin{aligned}[t]
    X &=  0.5Z_1 - 0.9Z_2 + 0.4Z_3^2 + Z_4Z_5\varepsilon_X, \\
  Y &= -0.8Z_4 + Z_5^2 + \exp(Z_6) + \sin(2\pi Z_7)\varepsilon_Y
    \end{aligned}
    $
    \item \label{eq:null3}$
    \begin{aligned}[t]
    X &=  \tanh(0.5Z_1 - 0.9Z_2 + Z_3 +\varepsilon_X), \\
  Y &= \exp(-0.8Z_4 Z_5  + 0.6Z_6Z_7+\varepsilon_Y)
    \end{aligned}
    $
    \item \label{eq:null4}$
    \begin{aligned}[t]
    X &= \sin(2\pi Z_1) + 0.1\varepsilon_X, \\
    Y &= \sin(2\pi Z_1)  + \varepsilon_Y
    \end{aligned}
    $
\end{enumerate}

The alternative settings include a small linear effect of $X$ on $Y$, a non-linear effect interacting with a covariate, and a non-linear effect on the variance. Settings \ref{eq:alt2} and \ref{eq:alt3} satisfy $\E_Z[\Cov(X,Y \mid Z)] = 0$ and setting \ref{eq:alt3} satisfies $\Cov(X,Y \mid Z) = 0$, which implies that the conditional dependencies in these two groups cannot be identified by the GCM and wGCM, respectively. 
The results are shown in Figure \ref{fig:alt}. 
\begin{enumerate}[label=\textbf{Alt. \arabic*}:,ref=\arabic*, align=left, labelwidth=!, leftmargin=*, itemsep = 0em]
    \item \label{eq:alt1}$
    \begin{aligned}[t]
    X &= 0.7Z_1 + Z_2 + \varepsilon_X, \\
    Y &= 0.4Z_3 - 0.2Z_4 - 0.1X  + \varepsilon_Y
    \end{aligned}
    $
    
    \item \label{eq:alt2}$
    \begin{aligned}[t]
    X &= \sin(Z_1) + \varepsilon_X, \\
    Y &= \tanh(Z_2) + 0.4 X^2 Z_3  + \varepsilon_Y
    \end{aligned}
    $
    \item \label{eq:alt3}$
    \begin{aligned}[t]
    X &= 0.2 Z_2^3 + \tanh(Z_4) + \varepsilon_X, \\
    Y &= \sin(\pi Z_1) - 0.4Z_2^2 + \cos(0.2 \pi X)\varepsilon_Y
    \end{aligned}
    $
\end{enumerate}

\paragraph{Results}

In the null settings \ref{eq:null1} -- \ref{eq:null3} GKCM RF has approximately nominal type-I error rates at each sample size, while all other tests have significantly inflated type-I error rates in multiple scenarios or sample sizes. 
In the challenging null setting \ref{eq:null4} by \citet{lundborg_projected_2024} GKCM RF approaches the nominal error rates with increasing sample size, while most other methods consistently reject the null hypothesis even at the highest sample size.

\begin{figure}[h]
    \centering
    \includegraphics[width=\linewidth]{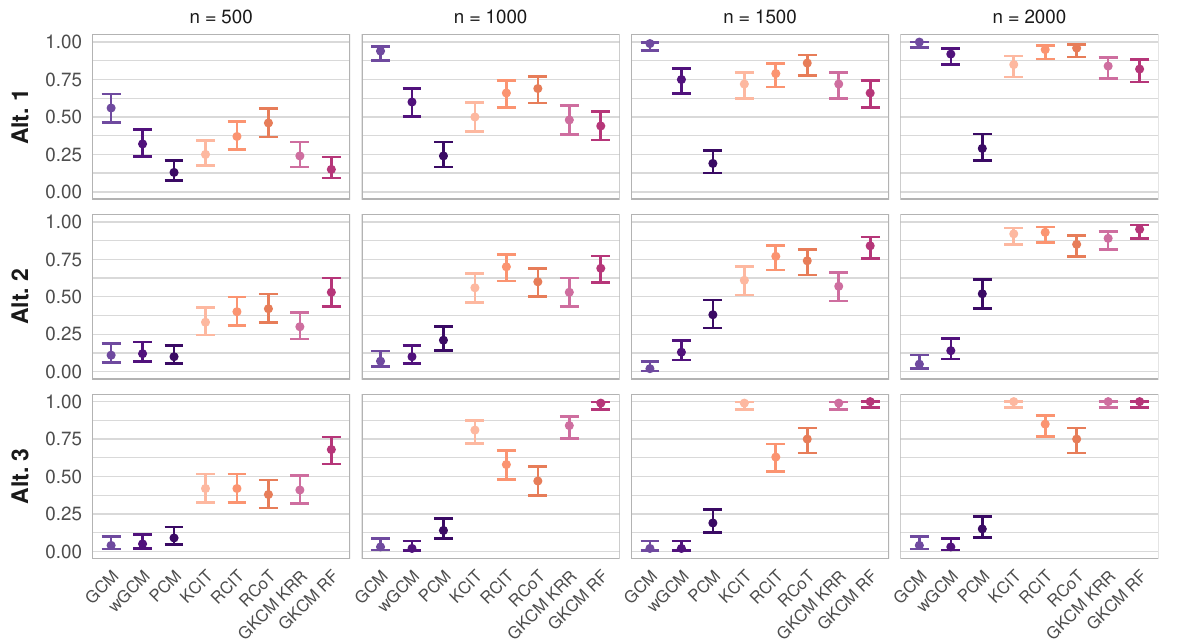}
    \label{fig:alt}
    \caption{Rejection rates in the alternative settings with rejection threshold $p < 0.05$ (100 iterations). Error bars indicate 95\% Wilson confidence intervals.}
\label{fig:alt}
\end{figure}

In the alternative setting \ref{eq:alt1} GKCM RF has below average power, yet remains competitive with the other kernel-based tests. KCIT and GKCM KRR show comparable power to GKCM RF despite having inflated type-I error rates in the linear null setting \ref{eq:null1}. In the nonlinear settings \ref{eq:alt2} and \ref{eq:alt3} GKCM RF has power greater than or equal to all other methods. GKCM RF thereby has considerable power under non-linear dependence and good level control across a diverse range of settings. These results indicate that random forests offer a promising alternative to KRR in kernel-based CI testing when hyperparameter tuning is not feasible. Additional simulations, including hyperparameter tuning, are provided in Appendix \ref{sec:addsim}. 

\section{Conclusion}

We introduced GKCM, established its theoretical properties and assessed its finite-sample performance using RKHS-valued random forests through a simulation study. GKCM frequently outperforms state-of-the-art alternatives, which shows its potential as a general CI test for statistical and causal inference.

However, several questions related to the practical use of GKCM remain open. Important directions include a systematic evaluation of additional regression methods, an investigation of whether computationally feasible hyperparameter tuning strategies can further improve performance, and an extension of the empirical study to settings with mixed data types. 

Furthermore, we provided sufficient conditions for uniform asymptotic type-I error guarantees. An additional direction of future work is to establish examples of families of distributions for which these conditions can be shown to hold.

\acks{LB was funded by the German Research Foundation (DFG) as part of the Research Unit “Lifespan AI: From Longitudinal Data to Lifespan Inference in Health” (DFG FOR 5347), Grant 459360854, and thanks Anton Rask Lundborg for helpful discussions. DS was supported in part by the Responsible AI Research Centre (RAIR).}

\bibliography{refs}

@book{anderson_introduction_2003,
	address = {Hoboken, New Jersey},
	edition = {Third},
	series = {Wiley series in probability and statistics},
	title = {An {Introduction} to {Multivariate} {Statistical} {Analysis}},
	isbn = {978-0-471-36091-9},
	language = {eng},
	publisher = {Wiley-Interscience},
	author = {Anderson, Theodore W.},
	year = {2003},
}

@article{shah_hardness_2020,
	title = {The {Hardness} of {Conditional} {Independence} {Testing} and the {Generalised} {Covariance} {Measure}},
	volume = {48},
	number = {3},
	journal = {The Annals of Statistics},
	author = {Shah, Rajen D. and Peters, Jonas},
	year = {2020},
	note = {Publisher: Institute of Mathematical Statistics},
	pages = {1514--1538},
}

@article{daudin_partial_1980,
	title = {Partial association measures and an application to qualitative regression},
	volume = {67},
	url = {https://doi.org/10.1093/biomet/67.3.581},
	number = {3},
	journal = {Biometrika},
	author = {Daudin, Jean-Jacques},
	year = {1980},
	pages = {581--590},
}

@article{scheidegger_weighted_2022,
	title = {The {Weighted} {Generalised} {Covariance} {Measure}},
	volume = {23},
	url = {http://jmlr.org/papers/v23/21-1328.html},
	number = {273},
	journal = {Journal of Machine Learning Research},
	author = {Scheidegger, Cyrill and Hörrmann, Julia and Bühlmann, Peter},
	year = {2022},
	pages = {1--68},
}

@article{dawid_conditional_1979,
	title = {Conditional {Independence} in {Statistical} {Theory}},
	volume = {41},
	copyright = {https://academic.oup.com/journals/pages/open\_access/funder\_policies/chorus/standard\_publication\_model},
	url = {https://academic.oup.com/jrsssb/article/41/1/1/7027599},
	language = {en},
	number = {1},
	urldate = {2024-10-17},
	journal = {Journal of the Royal Statistical Society Series B: Statistical Methodology},
	author = {Dawid, A. Philip},
	month = sep,
	year = {1979},
	pages = {1--15},
}

@article{strobl_approximate_2019,
	title = {Approximate {Kernel}-{Based} {Conditional} {Independence} {Tests} for {Fast} {Non}-{Parametric} {Causal} {Discovery}},
	volume = {7},
	copyright = {De Gruyter expressly reserves the right to use all content for commercial text and data mining within the meaning of Section 44b of the German Copyright Act.},
	url = {https://doi.org/10.1515/jci-2018-0017},
	language = {en},
	number = {1},
	urldate = {2024-11-10},
	journal = {Journal of Causal Inference},
	author = {Strobl, Eric V. and Zhang, Kun and Visweswaran, Shyam},
	month = mar,
	year = {2019},
	note = {Publisher: De Gruyter},
}

@inproceedings{zhang_kernel-based_2011,
	address = {Barcelona, Spain},
	series = {{UAI}'11},
	title = {Kernel-based {Conditional} {Independence} {Test} and {Application} in {Causal} {Discovery}},
	url = {https://webdav.tuebingen.mpg.de/causality/UAI11_KCItest.pdf},
	language = {en},
	booktitle = {Proceedings of the {Twenty}-{Seventh} {Conference} on {Uncertainty} in {Artificial} {Intelligence}},
	publisher = {AUAI Press},
	author = {Zhang, Kun and Peters, Jonas and Janzing, Dominik and Scholkopf, Bernhard},
	year = {2011},
	pages = {804--813},
}

@book{steinwart_support_2008,
	address = {New York, NY},
	series = {Information {Science} and {Statistics}},
	title = {Support {Vector} {Machines}},
	copyright = {https://www.springernature.com/gp/researchers/text-and-data-mining},
	isbn = {978-0-387-77241-7},
	url = {https://link.springer.com/10.1007/978-0-387-77242-4},
	language = {en},
	urldate = {2024-12-04},
	publisher = {Springer New York},
	author = {Steinwart, Ingo and Christmann, Andreas},
	year = {2008},
}

@article{glymour_review_2019,
	title = {Review of {Causal} {Discovery} {Methods} {Based} on {Graphical} {Models}},
	volume = {10},
	url = {https://www.frontiersin.org/journals/genetics/articles/10.3389/fgene.2019.00524/full},
	doi = {10.3389/fgene.2019.00524},
	language = {English},
	urldate = {2025-01-03},
	journal = {Frontiers in Genetics},
	author = {Glymour, Clark and Zhang, Kun and Spirtes, Peter},
	month = jun,
	year = {2019},
	note = {Publisher: Frontiers},
}

@misc{garreau_large_2018,
	title = {Large sample analysis of the median heuristic},
	url = {http://arxiv.org/abs/1707.07269},
	doi = {10.48550/arXiv.1707.07269},
	language = {en},
	urldate = {2025-01-10},
	publisher = {arXiv},
	author = {Garreau, Damien and Jitkrittum, Wittawat and Kanagawa, Motonobu},
	month = oct,
	year = {2018},
	note = {arXiv:1707.07269 [math]},
}

@book{berlinet_reproducing_2004,
	title = {Reproducing {Kernel} {Hilbert} {Spaces} {In} {Probability} and {Statistics}},
	isbn = {978-1-4020-7679-4},
	publisher = {Kluwer Academic Publishers},
	author = {Berlinet, Alain and Thomas-Agnan, Christine},
	year = {2004},
}

@article{lundborg_projected_2024,
	title = {The projected covariance measure for assumption-lean variable significance testing},
	volume = {52},
	url = {https://doi.org/10.1214/24-AOS2447},
	number = {6},
	urldate = {2025-02-17},
	journal = {The Annals of Statistics},
	author = {Lundborg, Anton Rask and Kim, Ilmun and Shah, Rajen D. and Samworth, Richard J.},
	month = dec,
	year = {2024},
	note = {Publisher: Institute of Mathematical Statistics},
	pages = {2851--2878},
}

@article{szabo_characteristic_2018,
	title = {Characteristic and {Universal} {Tensor} {Product} {Kernels}},
	volume = {18},
	url = {http://jmlr.org/papers/v18/17-492.html},
	number = {233},
	urldate = {2025-04-24},
	journal = {Journal of Machine Learning Research},
	author = {Szabó, Zoltán and Sriperumbudur, Bharath K.},
	year = {2018},
	pages = {1--29},
}

@inproceedings{grunewalder_conditional_2012,
	address = {Madison, WI, USA},
	series = {{ICML}'12},
	title = {Conditional mean embeddings as regressors},
	urldate = {2025-04-25},
	booktitle = {Proceedings of the 29th {International} {Coference} on {International} {Conference} on {Machine} {Learning}},
	publisher = {Omnipress},
	author = {Grünewälder, Steffen and Lever, Guy and Baldassarre, Luca and Patterson, Sam and Gretton, Arthur and Pontil, Massimilano},
	month = jun,
	year = {2012},
	pages = {1803--1810},
}

@inproceedings{rahimi_random_2007,
	title = {Random {Features} for {Large}-{Scale} {Kernel} {Machines}},
	volume = {20},
	url = {https://papers.nips.cc/paper_files/paper/2007/hash/013a006f03dbc5392effeb8f18fda755-Abstract.html},
	urldate = {2025-04-26},
	booktitle = {Advances in {Neural} {Information} {Processing} {Systems}},
	publisher = {Curran Associates, Inc.},
	author = {Rahimi, Ali and Recht, Benjamin},
	year = {2007},
}

@inproceedings{park_measure-theoretic_2020,
	title = {A {Measure}-{Theoretic} {Approach} to {Kernel} {Conditional} {Mean} {Embeddings}},
	volume = {33},
	url = {https://proceedings.neurips.cc/paper_files/paper/2020/file/f340f1b1f65b6df5b5e3f94d95b11daf-Paper.pdf},
	booktitle = {Advances in {Neural} {Information} {Processing} {Systems}},
	publisher = {Curran Associates, Inc.},
	author = {Park, Junhyung and Muandet, Krikamol},
	editor = {Larochelle, H. and Ranzato, M. and Hadsell, R. and Balcan, M. F. and Lin, H.},
	year = {2020},
	pages = {21247--21259},
}

@article{sriperumbudur_universality_2011,
	title = {Universality, {Characteristic} {Kernels} and {RKHS} {Embedding} of {Measures}},
	volume = {12},
	url = {http://jmlr.org/papers/v12/sriperumbudur11a.html},
	number = {70},
	urldate = {2025-05-22},
	journal = {Journal of Machine Learning Research},
	author = {Sriperumbudur, Bharath K. and Fukumizu, Kenji and Lanckriet, Gert R. G.},
	year = {2011},
	pages = {2389--2410},
}

@article{li_towards_2024,
	title = {Towards {Optimal} {Sobolev} {Norm} {Rates} for the {Vector}-{Valued} {Regularized} {Least}-{Squares} {Algorithm}},
	volume = {25},
	url = {https://www.jmlr.org/papers/volume25/23-1663/23-1663.pdf},
	language = {en},
	number = {1},
	journal = {J. Mach. Learn. Res.},
	author = {Li, Zhu and Meunier, Dimitri and Mollenhauer, Mattes and Gretton, Arthur},
	year = {2024},
	pages = {8554--8604},
}

@article{lundborg_conditional_2022,
	title = {Conditional independence testing in {Hilbert} spaces with applications to functional data analysis},
	volume = {84},
	url = {https://doi.org/10.1111/rssb.12544},
	language = {en},
	number = {5},
	urldate = {2025-06-05},
	journal = {Journal of the Royal Statistical Society: Series B (Statistical Methodology)},
	author = {Lundborg, Anton Rask and Shah, Rajen D. and Peters, Jonas},
	year = {2022},
	pages = {1821--1850},
}

@inproceedings{li_optimal_2022,
	address = {Red Hook, NY, USA},
	series = {{NIPS} '22},
	title = {Optimal {Learning} {Rates} for {Regularized} {Conditional} {Mean} {Embedding}},
	url = {https://doi.org/10.52202/068431-0320},
	urldate = {2025-06-06},
	booktitle = {Proceedings of the 36th {International} {Conference} on {Neural} {Information} {Processing} {Systems}},
	publisher = {Curran Associates Inc.},
	author = {Li, Zhu and Meunier, Dimitri and Mollenhauer, Mattes and Gretton, Arthur},
	month = nov,
	year = {2022},
	pages = {4433--4445},
}

@inproceedings{geurts_kernelizing_2006,
	address = {Pittsburgh, Pennsylvania},
	series = {{ICML}'06},
	title = {Kernelizing the {Output} of {Tree}-{Based} {Methods}},
	url = {https://doi.org/10.1145/1143844.1143888},
	language = {en},
	urldate = {2025-06-21},
	booktitle = {Proceedings of the 23nd {International} {Machine} {Learning} {Conference}},
	publisher = {Association for Computing Machinery},
	author = {Geurts, Pierre and Wehenkel, Louis and d'Alché-Buc, Florence},
	year = {2006},
	pages = {345--352},
}

@inproceedings{geurts_gradient_2007,
	address = {New York, NY, USA},
	series = {{ICML}'07},
	title = {Gradient {Boosting} for {Kernelized} {Output} {Spaces}},
	url = {https://doi.org/10.1145/1273496.1273533},
	doi = {10.1145/1273496.1273533},
	urldate = {2025-06-26},
	booktitle = {Proceedings of the 24th international conference on {Machine} learning},
	publisher = {Association for Computing Machinery},
	author = {Geurts, Pierre and Wehenkel, Louis and d'Alché-Buc, Florence},
	month = jun,
	year = {2007},
	pages = {289--296},
}

@article{cevid_distributional_2022,
	title = {Distributional {Random} {Forests}: {Heterogeneity} {Adjustment} and {Multivariate} {Distributional} {Regression}},
	volume = {23},
	shorttitle = {Distributional {Random} {Forests}},
	url = {http://jmlr.org/papers/v23/21-0585.html},
	number = {333},
	urldate = {2025-07-02},
	journal = {Journal of Machine Learning Research},
	author = {Cevid, Domagoj and Michel, Loris and Näf, Jeffrey and Bühlmann, Peter and Meinshausen, Nicolai},
	year = {2022},
	pages = {1--79},
}

@book{paulsen_introduction_2016,
	address = {Cambridge},
	series = {Cambridge {Studies} in {Advanced} {Mathematics}},
	title = {An {Introduction} to the {Theory} of {Reproducing} {Kernel} {Hilbert} {Spaces}},
	isbn = {978-1-107-10409-9},
	url = {https://doi.org/10.1017/CBO9781316219232},
	urldate = {2025-08-08},
	publisher = {Cambridge University Press},
	author = {Paulsen, Vern I. and Raghupathi, Mrinal},
	year = {2016},
}

@article{bodenham_comparison_2016,
	title = {A comparison of efficient approximations for a weighted sum of chi-squared random variables},
	volume = {26},
	url = {https://doi.org/10.1007/s11222-015-9583-4},
	language = {en},
	number = {4},
	urldate = {2025-08-29},
	journal = {Statistics and Computing},
	author = {Bodenham, Dean A. and Adams, Niall M.},
	month = jul,
	year = {2016},
	pages = {917--928},
}

@article{constantinou_extended_2017,
	title = {Extended conditional independence and applications in causal inference},
	volume = {45},
	url = {https://doi.org/10.1214/16-AOS1537},
	number = {6},
	urldate = {2025-09-02},
	journal = {The Annals of Statistics},
	author = {Constantinou, Panayiota and Dawid, A. Philip},
	month = dec,
	year = {2017},
	note = {Publisher: Institute of Mathematical Statistics},
	pages = {2618--2653},
}

@book{cohn_measure_2013,
	address = {New York, NY},
	series = {Birkhäuser {Advanced} {Texts} {Basler} {Lehrbücher}},
	title = {Measure {Theory}: {Second} {Edition}},
	isbn = {978-1-4614-6955-1},
	shorttitle = {Measure {Theory}},
	url = {https://link.springer.com/10.1007/978-1-4614-6956-8},
	language = {en},
	urldate = {2025-10-28},
	publisher = {Springer},
	author = {Cohn, Donald L.},
	year = {2013},
}

@book{kallenberg_foundations_2021,
	address = {Cham},
	series = {Probability {Theory} and {Stochastic} {Modelling}},
	title = {Foundations of {Modern} {Probability}},
	volume = {99},
	isbn = {978-3-030-61871-1},
	url = {https://link.springer.com/10.1007/978-3-030-61871-1},
	language = {en},
	urldate = {2025-11-06},
	publisher = {Springer International Publishing},
	author = {Kallenberg, Olav},
	year = {2021},
}

@book{hsing_theoretical_2015,
	title = {Theoretical {Foundations} of {Functional} {Data} {Analysis}, with an {Introduction} to {Linear} {Operators}},
	isbn = {978-1-118-76257-8},
	url = {https://onlinelibrary.wiley.com/doi/book/10.1002/9781118762547},
	publisher = {Wiley},
	author = {Hsing, Tailen and Eubank, Randall L.},
	year = {2015},
}

@inproceedings{zhang_feature--feature_2017,
	address = {Sydney, Australia},
	title = {Feature-to-{Feature} {Regression} for a {Two}-{Step} {Conditional} {Independence} {Test}},
	volume = {33},
	url = {https://www.auai.org/uai2017/proceedings/papers/250.pdf},
	urldate = {2025-12-02},
	booktitle = {Proceedings of the {Thirty}-{Third} {Conference} on {Uncertainty} in {Artificial} {Intelligence}},
	publisher = {AUAI Press},
	author = {Zhang, Qinyi and Filippi, Sarah and Flaxman, Seth and Sejdinovic, Dino},
	year = {2017},
}

@inproceedings{mastouri_proximal_2021,
	title = {Proximal {Causal} {Learning} with {Kernels}: {Two}-{Stage} {Estimation} and {Moment} {Restriction}},
	shorttitle = {Proximal {Causal} {Learning} with {Kernels}},
	url = {https://proceedings.mlr.press/v139/mastouri21a.html},
	language = {en},
	urldate = {2025-12-03},
	booktitle = {Proceedings of the 38th {International} {Conference} on {Machine} {Learning}},
	publisher = {PMLR},
	author = {Mastouri, Afsaneh and Zhu, Yuchen and Gultchin, Limor and Korba, Anna and Silva, Ricardo and Kusner, Matt and Gretton, Arthur and Muandet, Krikamol},
	month = jul,
	year = {2021},
	pages = {7512--7523},
}

@inproceedings{zhang_identifiability_2009,
	address = {Arlington, Virginia, USA},
	series = {{UAI} '09},
	title = {On the identifiability of the post-nonlinear causal model},
	url = {https://dl.acm.org/doi/10.5555/1795114.1795190},
	urldate = {2025-12-05},
	booktitle = {Proceedings of the {Twenty}-{Fifth} {Conference} on {Uncertainty} in {Artificial} {Intelligence}},
	publisher = {AUAI Press},
	author = {Zhang, Kun and Hyvärinen, Aapo},
	month = jun,
	year = {2009},
	pages = {647--655},
}

@inproceedings{el_ahmad_deep_2024,
	address = {Cham},
	title = {Deep {Sketched} {Output} {Kernel} {Regression} for {Structured} {Prediction}},
	url = {https://doi.org/10.1007/978-3-031-70352-2_6},
	language = {en},
	booktitle = {Machine {Learning} and {Knowledge} {Discovery} in {Databases}. {Research} {Track}},
	publisher = {Springer Nature Switzerland},
	author = {El Ahmad, Tamim and Yang, Junjie and Laforgue, Pierre and d’Alché-Buc, Florence},
	editor = {Bifet, Albert and Davis, Jesse and Krilavičius, Tomas and Kull, Meelis and Ntoutsi, Eirini and Žliobaitė, Indrė},
	year = {2024},
	pages = {93--110},
}

@article{peters_causal_2016,
	title = {Causal {Inference} by using {Invariant} {Prediction}: {Identification} and {Confidence} {Intervals}},
	volume = {78},
	shorttitle = {Causal {Inference} by using {Invariant} {Prediction}},
	url = {https://doi.org/10.1111/rssb.12167},
	number = {5},
	urldate = {2025-12-19},
	journal = {Journal of the Royal Statistical Society Series B: Statistical Methodology},
	author = {Peters, Jonas and Bühlmann, Peter and Meinshausen, Nicolai},
	month = nov,
	year = {2016},
	pages = {947--1012},
}

@article{heinze-deml_invariant_2018,
	title = {Invariant {Causal} {Prediction} for {Nonlinear} {Models}},
	volume = {6},
	url = {https://doi.org/10.1515/jci-2017-0016},
	number = {2},
	journal = {Journal of Causal Inference},
	author = {Heinze-Deml, Christina and Peters, Jonas and Meinshausen, Nicolai},
	month = sep,
	year = {2018},
}

@article{brouard_input_2016,
	title = {Input {Output} {Kernel} {Regression}: {Supervised} and {Semi}-{Supervised} {Structured} {Output} {Prediction} with {Operator}-{Valued} {Kernels}},
	volume = {17},
	issn = {1533-7928},
	shorttitle = {Input {Output} {Kernel} {Regression}},
	url = {http://jmlr.org/papers/v17/15-602.html},
	number = {176},
	urldate = {2025-12-22},
	journal = {Journal of Machine Learning Research},
	author = {Brouard, Céline and Szafranski, Marie and d'Alché-Buc, Florence},
	year = {2016},
	pages = {1--48},
}

@inproceedings{laforgue_duality_2020,
	title = {Duality in {RKHSs} with {Infinite} {Dimensional} {Outputs}: {Application} to {Robust} {Losses}},
	shorttitle = {Duality in {RKHSs} with {Infinite} {Dimensional} {Outputs}},
	url = {https://proceedings.mlr.press/v119/laforgue20a.html},
	language = {en},
	urldate = {2025-12-22},
	booktitle = {Proceedings of the 37th {International} {Conference} on {Machine} {Learning}},
	publisher = {PMLR},
	author = {Laforgue, Pierre and Lambert, Alex and Brogat-Motte, Luc and D’Alché-Buc, Florence},
	month = nov,
	year = {2020},
	pages = {5598--5607},
}

@inproceedings{shimizu_neural-kernel_2024,
	address = {Vienna, Austria},
	series = {{ICML}'24},
	title = {Neural-kernel conditional mean embeddings},
	volume = {235},
	url = {https://dl.acm.org/doi/10.5555/3692070.3693903},
	urldate = {2025-12-22},
	booktitle = {Proceedings of the 41st {International} {Conference} on {Machine} {Learning}},
	publisher = {JMLR.org},
	author = {Shimizu, Eiki and Fukumizu, Kenji and Sejdinovic, Dino},
	month = jul,
	year = {2024},
	pages = {45040--45059},
}

@misc{zhang_inclusion_2011,
	title = {On the {Inclusion} {Relation} of {Reproducing} {Kernel} {Hilbert} {Spaces}},
	url = {http://arxiv.org/abs/1106.4075},
	doi = {10.48550/arXiv.1106.4075},
	language = {en},
	urldate = {2026-03-12},
	publisher = {arXiv},
	author = {Zhang, Haizhang and Zhao, Liang},
	month = jun,
	year = {2011},
	note = {arXiv:1106.4075 [math]},
}

@inproceedings{he_hardness_2025,
	address = {San Diego, California},
	title = {On the {Hardness} of {Conditional} {Independence} {Testing} {In} {Practice}},
	language = {en},
	urldate = {2026-03-20},
	booktitle = {Advances in {Neural} {Information} {Processing} {Systems} 39},
	publisher = {NeurIPS},
	author = {He, Zheng and Pogodin, Roman and Li, Yazhe and Deka, Namrata and Gretton, Arthur and Sutherland, Danica J.},
	year = {2025},
}

@misc{pogodin_practical_2025,
	title = {Practical {Kernel} {Tests} of {Conditional} {Independence}},
	url = {http://arxiv.org/abs/2402.13196},
	doi = {10.48550/arXiv.2402.13196},
	language = {en},
	urldate = {2026-03-20},
	publisher = {arXiv},
	author = {Pogodin, Roman and Schrab, Antonin and Li, Yazhe and Sutherland, Danica J. and Gretton, Arthur},
	month = sep,
	year = {2025},
	note = {arXiv:2402.13196 [cs]},
}

@inproceedings{meunier_optimal_2024,
	address = {Red Hook, NY, USA},
	series = {{NIPS} '24},
	title = {Optimal {Rates} for {Vector}-{Valued} {Spectral} {Regularization} {Learning} {Algorithms}},
	isbn = {9798331314385},
	urldate = {2026-04-02},
	booktitle = {Proceedings of the 38th {International} {Conference} on {Neural} {Information} {Processing} {Systems}},
	publisher = {Curran Associates Inc.},
	author = {Meunier, Dimitri and Shen, Zikai and Mollenhauer, Mattes and Gretton, Arthur and Li, Zhu},
	year = {2024},
	pages = {82514--82559},
}

@Manual{kook_comets_2025,
    title = {comets: Covariance Measure Tests for Conditional Independence},
    author = {Lucas Kook},
    year = {2025},
    note = {R package version 0.2-2},
    url = {https://CRAN.R-project.org/package=comets},
    doi = {10.32614/CRAN.package.comets},
  }

@Manual{michel_drf_2021,
    title = {drf: Distributional Random Forests},
    author = {Loris Michel and Domagoj Cevid},
    year = {2021},
    note = {R package version 1.1.0},
    url = {https://CRAN.R-project.org/package=drf},
    doi = {10.32614/CRAN.package.drf},
  }

@book{da_prato_stochastic_2014,
	address = {Cambridge},
	edition = {2},
	series = {Encyclopedia of {Mathematics} and its {Applications}},
	title = {Stochastic {Equations} in {Infinite} {Dimensions}},
	isbn = {978-1-107-05584-1},
	url = {https://doi.org/10.1017/CBO9781107295513},
	urldate = {2026-04-03},
	publisher = {Cambridge University Press},
	author = {Da Prato, Giuseppe and Zabczyk, Jerzy},
	year = {2014},
}

\appendix

\section{Proof of Lemma \ref{lem:sigequal}} \label{sec:proof}

By \citet[Lemma 8.3.8]{cohn_measure_2013}, there exists a Borel-measurable function $g: \F \to \X$ such that $(g \circ\phi)(x) = x$ for each $x \in \X$. Thereby $X = (g \circ \phi)(X)$ and $\sigma(X) = \sigma((g \circ\phi)(X))$. Since $g$ and $\phi$ are Borel-measurable, it holds that $\sigma((g \circ\phi)(X)) = \sigma(X) \subseteq\sigma(\phi(X))$ and $\sigma(\phi(X)) \subseteq \sigma(X)$. 

\section{CI testing with joint embeddings} \label{sec:joint}

In the following, we review the use of joint embeddings in kernel-based CI testing. Two different approaches have been proposed. The first regresses the joint embedding of $(X,Z)$ directly on $Z$. The second exploits a decomposition in which only the embedding of $X$ is regressed on $Z$, while the $Z$-embedding is incorporated afterwards. Although the latter avoids some of the difficulties of the former, both methods encounter distinct statistical and practical issues, which we discuss below.

The direct approach is used in the original definition of the KCIT \citep{zhang_kernel-based_2011}.
To construct the joint embedding, \citeauthor{zhang_kernel-based_2011} combine the Gaussian kernel $k$ with a (possibly differently scaled) Gaussian kernel $m': \Z \times \Z\to \R$ with RKHS ${\Hc}'$ and canonical feature map $\psi'(z) :=  m'(\cdot, z)$. They use the Gaussian tensor-product kernel 
\begin{equation*}
    k'\big((x,z),(x',z')\big) := k(x,x')m'(z,z') 
\end{equation*}
whose RKHS is the completed Hilbert tensor product $\F \otimes {\Hc}'$ with canonical feature map 
\begin{equation*}
    \phi'(x,z) := \phi(x) \otimes \psi'(z)
\end{equation*}
\citep[][Section 5.5]{paulsen_introduction_2016}. 

However, direct regression of $\phi'(X,Z)$ is generally problematic: Not only is the output space large, but the regression function additionally has the form $z \mapsto F_P(z) \otimes \psi'(z)$. 
The regression method thereby needs to simultaneously learn $z \mapsto F_P(z)$ and reconstruct $z \mapsto \psi'(z)$, even though the latter is deterministic and fully known. Furthermore, the latter part of the problem behaves like learning an identity operator between infinite-dimensional RKHSs, which is typically not Hilbert-Schmidt \citep[][Appendix B.2]{pogodin_practical_2025}. As a consequence, the regression function is misaligned with the hypothesis spaces of KRR (cf. Section \ref{sec:KRR}). This phenomenon is illustrated by \citet[][Appendix B.9]{mastouri_proximal_2021}, who show that regularisation leads to considerable shrinkage especially in data-sparse regions. The resulting bias is an important source of Type-I error inflation in the original KCIT. 

To avoid these difficulties, \citet{pogodin_practical_2025} propose to modify KCIT by regressing the joint embedding indirectly. Since $\psi'(Z)$ is $\sigma(Z)$-measurable, it can be pulled out of the conditional mean embedding:
\begin{equation*}
    \E[\phi'(X, Z) \mid Z] = \E[\phi(X) \otimes  \psi'(Z) \mid Z] = \E[\phi(X) \mid Z] \otimes  \psi'(Z)
\end{equation*}
By the bilinearity of the tensor product map, the residual then factorises as
\begin{equation*}
    \phi'(X,Z) - \E[\phi'(X,Z) \mid Z] =  \big(\phi(X) - \E[\phi(X) \mid Z]\big) \otimes \psi'(Z).
\end{equation*}
Accordingly, one estimates the $\F$-valued residual $\phi(X) - \E[\phi(X) \mid Z]$ and subsequently tensors it with $\psi'(Z)$. 
This removes the identity-like component from the regression and thereby eliminates a major source of type-I error inflation in the original KCIT.

Additionally, the decomposition has the benefit that the regression-rate requirements of the GHCM framework are not strengthened by using the joint embedding. If we define the joint conditional mean embedding function $F'_P (z) := F_P(z) \otimes \psi'(z)$ and its estimate by 
$\hat{F}'_n(z) := \hat{F}_n(z) \otimes \psi'(z)$, then the in-sample MSPE 
\begin{equation*}
    \mathcal{E}_n^{F'} := \frac{1}{n} \sum_{i=1}^n \lVert F'_P(Z_i) - \hat{F}'_n(Z_i) \rVert_{\mathcal{F} \otimes {\Hc}'}^2
\end{equation*}
satisfies
\begin{equation*}
    \mathcal{E}_n^{F'} = \frac{1}{n} \sum_{i=1}^n \lVert (F_P(Z_i) - \hat{F}_n(Z_i) ) \otimes  \psi'(Z_i) \rVert_{\mathcal{F} \otimes {\Hc}'}^2 
    =\frac{1}{n} \sum_{i=1}^n \lVert F_P(Z_i) - \hat{F}_n(Z_i) \rVert_{\mathcal{F}}^2 m'(Z_i, Z_i),
\end{equation*}
since $\lVert f \otimes h' \rVert_{\F \otimes {\Hc}'}^2 = \lVert f \rVert_{\F}^2 \lVert h' \rVert_{{\Hc}'}^2$ for simple tensors and $\lVert \psi'(z) \rVert_{{\Hc}'}^2 = m'(z,z)$. For a Gaussian kernel normalised so that $m'(z, z) = 1$, $\mathcal{E}_n^{F'}$ and $\mathcal{E}_n^{F}$ thereby coincide.

The decomposed joint embedding does, however, modify the test statistic through off-diagonal weights, and may thereby introduce a new source of finite-sample type-I errors. If we define the (uncentered) joint residuals by $\hat{\varepsilon}'_i := (\phi(X_i) - \hat{F}_n(Z_i)) \otimes \psi'(Z_i)$, then
\begin{equation*}
    \langle \hat{\varepsilon}'_i, \hat{\varepsilon}'_j\rangle_{\mathcal{F} \otimes {\Hc}'} = m'(Z_i, Z_j)\langle \phi(X_i) - \hat{F}_n(Z_i), \phi(X_j) - \hat{F}_n(Z_j) \rangle_{\F}, 
\end{equation*}
so replacing $\hat{\varepsilon}_i$ by $\hat{\varepsilon}'_i$ is equivalent to inserting the kernel weights $m'(Z_i, Z_j)$ and the corresponding centering corrections into the quadratic form defining the test statistic $T_n$. This reweighting does not change the form of the null distribution asymptotically when the regression errors satisfy the in-sample error conditions. In finite samples, however, there are typically non-negligible regression errors, and the weights can make their effect on the test statistic more pronounced. This problem is highlighted theoretically and empirically by \citet[][Section 5]{he_hardness_2025} for a version of the KCIT based on a U-statistic and the wild bootstrap: they show that using a joint embedding can improve power under the alternative, but also amplify ``leaked dependence'' under the null. This phenomenon may be especially pronounced when tuning $m'$ to maximise power. While leaked dependence should become negligible as the regression accuracy improves with growing sample size, \citeauthor{he_hardness_2025} demonstrate that it can remain practically relevant at low and moderate sample sizes at which kernel-based CI tests are typically used. Although \citeauthor{he_hardness_2025} only establish this phenomenon for tests based on U-statistics, the underlying mechanism also suggests potential finite-sample distortions for GHCM-style tests.

\section{Additional simulation results} \label{sec:addsim}

In the following, we include additional simulations. In Section \ref{sec:KCITsim}, we replicate the simulation study of \citet{zhang_kernel-based_2011} using the methods and hyperparameter settings as described in the main paper. In Section \ref{sec:newparsim}, we repeat the simulation from the main paper with alternative hyperparameter settings for the regression methods of the tests we compare to GKCM RF. 

\subsection{Simulation study from \citet{zhang_kernel-based_2011}} \label{sec:KCITsim}

The simulation study comprises four scenarios (cases I and II, each under the null and alternative) and considers varying sample sizes ($n \in \{200,400\}$) and numbers of conditioning variables ($d \in \{1, \dots, 5\}$). These four data-generating processes differ along two dimensions. First, cases I and II differ in how $X$ and $Y$ depend on the covariates $Z$, in particular in how many components of $Z$ are relevant. In case I, both $X$ and $Y$ depend only on $Z_1$, while the remaining covariates are irrelevant. In case II, $X$ and $Y$ depend on all components of $Z = (Z_1, \dots, Z_d)$. Second, the null and the alternative differ in whether there is an unobserved additive common cause of $X$ and $Y$: under the null, no latent common cause is added, whereas under the alternative, an additional latent common cause is added to both variables.

Let $(\varepsilon_X, \varepsilon_Y) \sim \mathcal{N}_2(0,\mathrm{I}_2)$, $Z = (Z_1, \dots, Z_d)\sim \mathcal{N}_d(0,\mathrm{I}_d)$, and $C \sim \mathcal{N}(0,0.25)$ be mutually independent. For each fixed scenario, the random vectors are independent across observations and the Monte Carlo iterations are independent of each other. However, within a given iteration, the datasets generated for different values of $d$ are not independent, since those for larger $d$ reuse and extend quantities generated for smaller $d$. To simplify notation, define
\begin{equation*}
    \phi(u) = u+\frac{u^3}{3}+\frac12\tanh\!\left(\frac{u}{3}\right),\qquad
    \psi(v) = v+\tanh\!\left(\frac{v}{3}\right),\qquad
    h(u) = \frac{u}{2}+0.7\tanh(u).
\end{equation*}
Additionally, let
\begin{equation*}
    f_1(z) = 0.7\left(\frac{z_1^3}{5} + \frac{z_1}{2}\right),\qquad
    g_1(z) = \frac{z_1^3/4+z_1}{3}.
\end{equation*}
For $j=2,\dots,d$, define recursively
\begin{equation*}
    f_j(z) = h\!\left(a_j f_{j-1}(z)+b_j z_j\right),\qquad
    g_j(z) = h\!\left(a_j g_{j-1}(z)+b_j z_j\right),
\end{equation*}
where
\begin{equation*}
    a_j=
    \begin{cases}
        \frac12,& j=2,\\[2mm]
        \frac23,& j\ge 3,
    \end{cases}
    \qquad
    b_j=
    \begin{cases}
        1,& j=2,\\[2mm]
        \frac56,& j\ge 3.
    \end{cases}
\end{equation*}
The scenarios are defined as follows.

\paragraph{Null and alternative.}
Under the null
\begin{equation*}
    X = \phi\bigl(f(Z)+\tanh(\varepsilon_X)\bigr),\qquad 
    Y = \psi\bigl(g(Z)+\varepsilon_Y\bigr).
\end{equation*}
Under the alternative
\begin{equation*}
    X=\phi\bigl(f(Z)+\tanh(\varepsilon_X)\bigr)+C,\qquad
    Y=\psi\bigl(g(Z)+\varepsilon_Y\bigr)+C.
\end{equation*}

\paragraph{Cases I and II.}
In case I, the functions $f(Z)$ and $g(Z)$ depend only on the first covariate:
\begin{equation*}
    f(Z)=f_1(Z),\qquad g(Z)=g_1(Z).
\end{equation*}
In case II, the functions are built recursively from all $d$ covariates:
\begin{equation*}
    f(Z)=f_d(Z),\qquad g(Z)=g_d(Z).
\end{equation*}

While these equations describe the structural form of the data-generating process, in the implemented simulation $X$, $Y$, and $Z$ are standardised before adding the common cause $C$ under the alternative.

\begin{figure}[p]
    \centering
    \includegraphics[width=\linewidth]{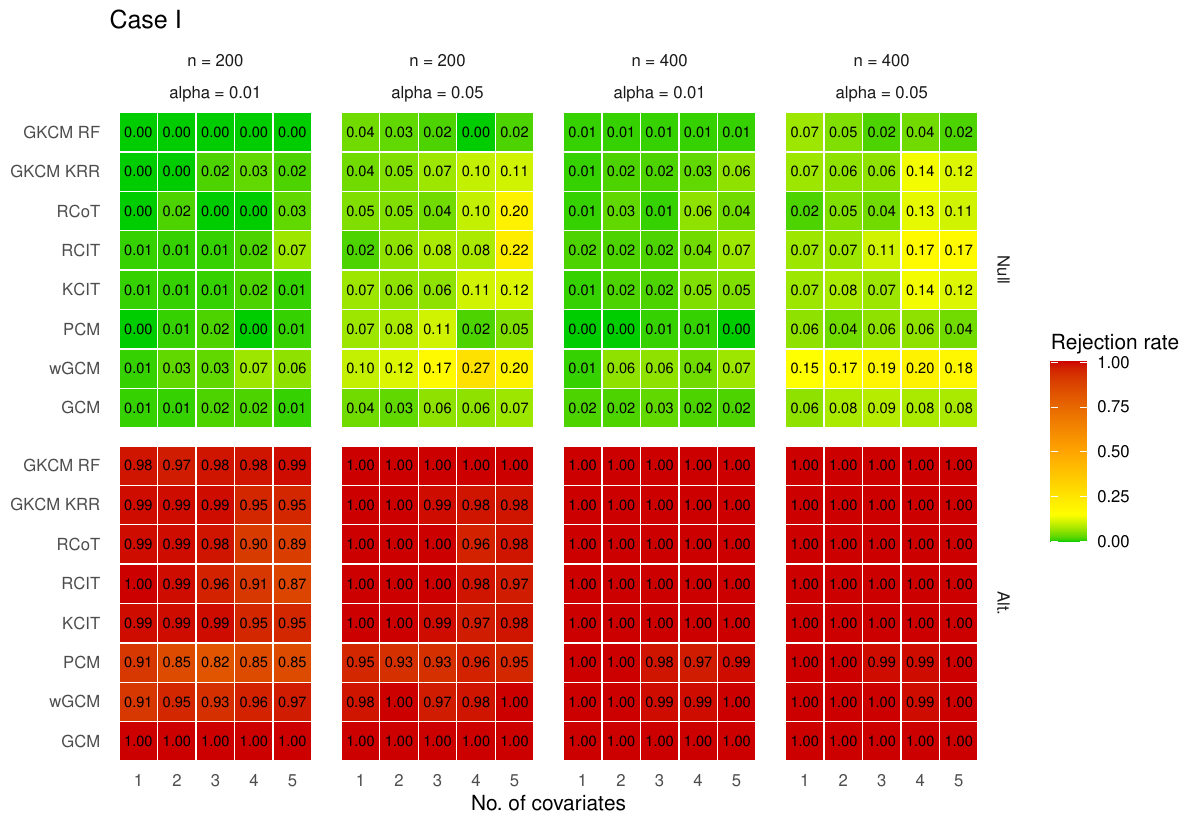}
    \includegraphics[width=\linewidth]{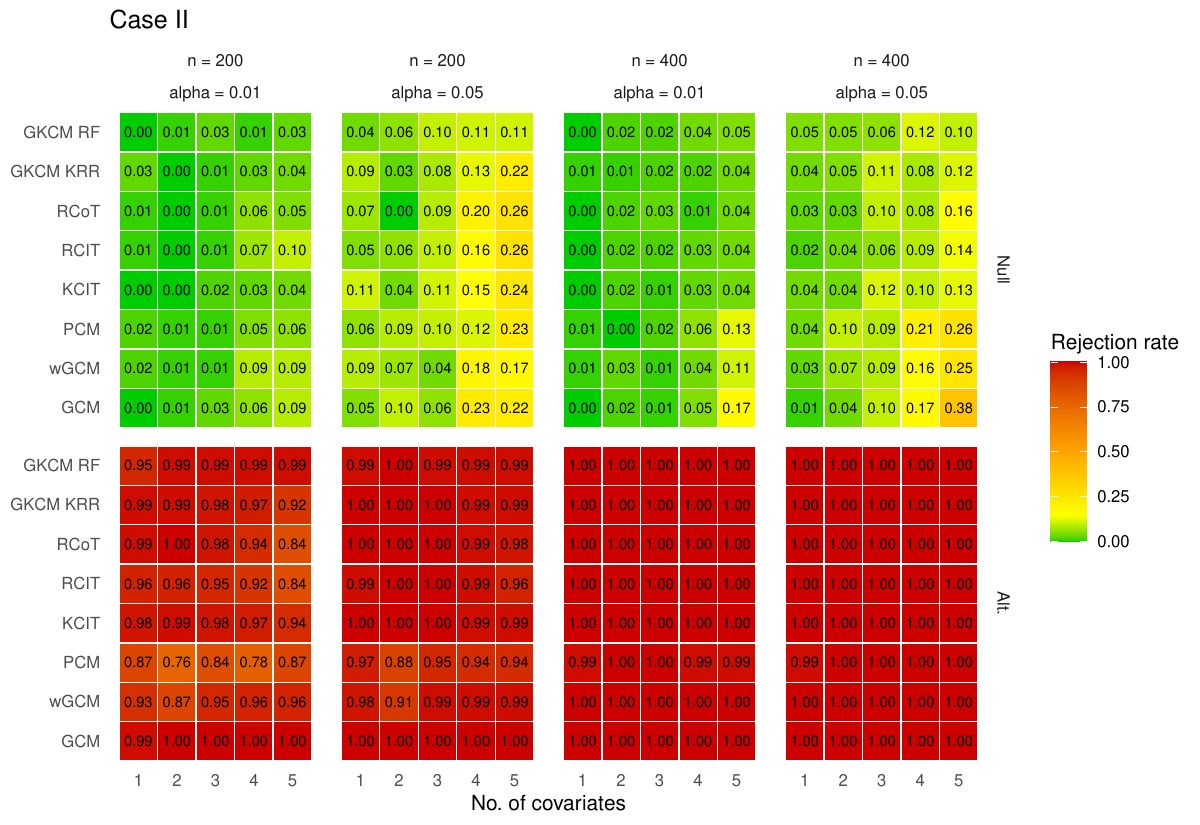}
    \label{fig:kcit_sim}
    \captionsetup{skip=-11pt}
    \caption{Rejection rates in the scenarios by \citet{zhang_kernel-based_2011} (100 iterations).}
\label{fig:kcit_sim}
\end{figure}

The results are shown in Figure \ref{fig:kcit_sim}. In case I, GKCM RF has excellent power and type I error rate, almost consistently outperforming the other methods. In case II, like the other tests, GKCM RF has slightly inflated type-I error rates for larger conditioning sets, yet is still able to outperform most other methods in many settings both with regard to level and power. The additional simulation thereby confirms the good performance of GKCM RF reported in the main text.

\subsection{New hyperparameter settings} \label{sec:newparsim}

In order to investigate the impact of the hyperparameter settings on the regression methods, we repeat the simulation study from the main paper with alternative, potentially improved settings for the regression methods of the competing tests. For all kernel-based methods using kernel ridge regression we used the same settings as in the main paper except for the regularisation parameter $\lambda$, which we tuned via leave-one-out cross-validation over the set $\{10^{-5}/n, ..., 10^3/n\}$ in each iteration (using a subsample of maximum size $1000$). The aim is to compare the RKHS-valued regression methods for settings in which light parameter tuning for kernel ridge regression is feasible. For the random forests employed in the residual-based methods, we use the same fixed parameter values as in the Distributional Random Forests (i.e., \texttt{num.trees} = 700, \texttt{mtry} = 7, and \texttt{min.node.size} = 5) in order to investigate the impact of the auto-tuning on their performance (note that these settings differ from the default fixed hyperparameter-settings used in the \texttt{comets} package).

\begin{figure}[h]
    \centering
    \includegraphics[width=\linewidth]{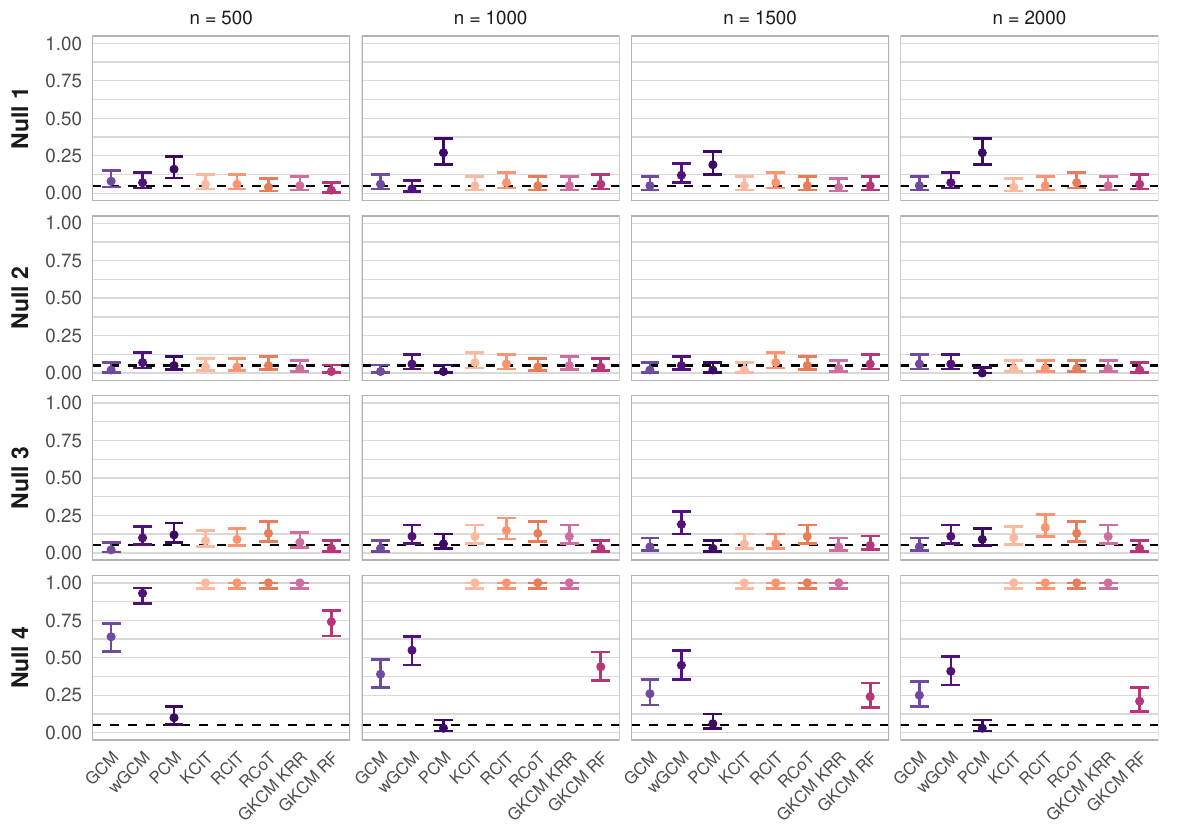}
    \caption{Rejection rates in the null settings with rejection threshold $p < 0.05$ (100 iterations). Error bars indicate 95\% Wilson confidence intervals and dashed lines the nominal level.}
\label{fig:null2}
\end{figure}

The results for the null settings are shown in Figure \ref{fig:null2}. For the kernel-based methods using KRR, tuning $\lambda$ notably improved the regressions in settings \ref{eq:null1} and \ref{eq:null2}, resulting in approximately nominal type-I error rates for all methods and sample sizes. The type-I error rates also improved in setting \ref{eq:null3}, yet remain mildly inflated in some sample sizes. Only in setting \ref{eq:null4} the performance did not improve. 

The performance of the residual-based methods improved as well, especially for PCM, which had type-I error rates exceeding 0.5 throughout settings \ref{eq:null1}, \ref{eq:null3}, and \ref{eq:null4} in the main text. Particularly in setting \ref{eq:null4}, PCM now stands out with approximately nominal type-I error rates for all sample sizes, thereby outperforming all other methods. For GCM and wGCM, the type-I error rates have improved in settings \ref{eq:null1} and \ref{eq:null4}. 

\begin{figure}[h]
    \centering
    \includegraphics[width=\linewidth]{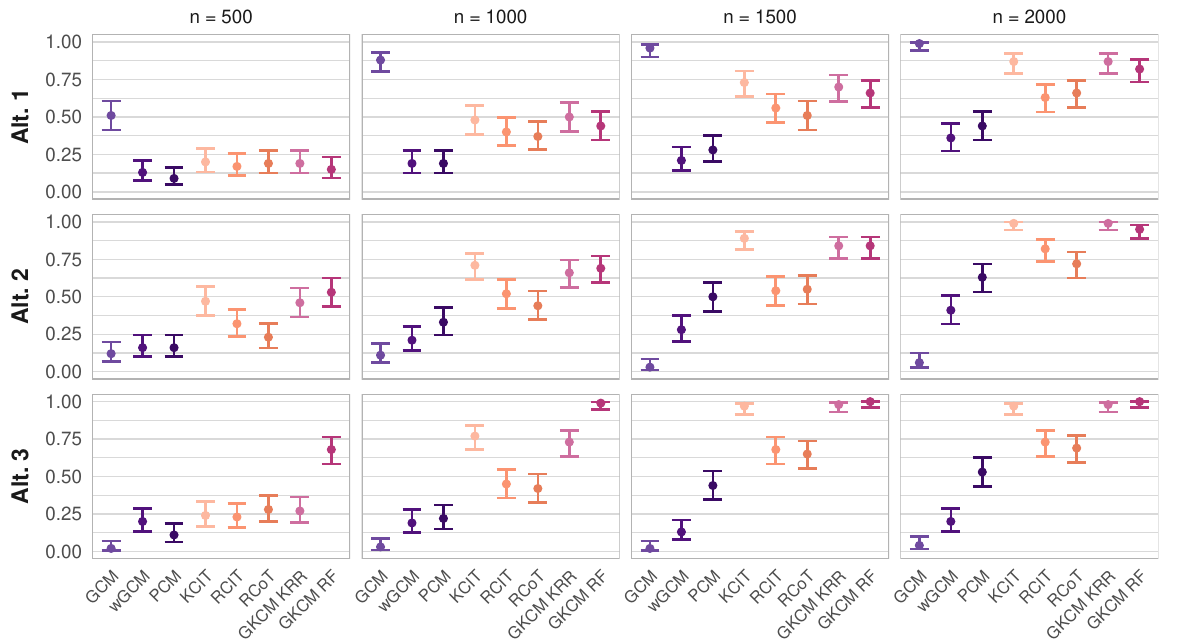}
    \label{fig:alt2}
    \caption{Rejection rates in the alternative settings with rejection threshold $p < 0.05$ (100 iterations). Error bars indicate 95\% Wilson confidence intervals.}
\label{fig:alt2}

\end{figure}

The results for the alternative settings are shown in Figure \ref{fig:alt2}. Like the type-I error rates, the power of the kernel-based tests using KRR has become more similar to the power of GKCM RF as well; hence, for the former, the reduction in type-I error rates came at the price of a reduction in power. In particular, in settings \ref{eq:alt1} and \ref{eq:alt2}, KCIT, GKCM KRR and GKCM RF now perform comparably (for most sample sizes with a small lead for the former). While RCIT and RCoT had superior power in the main paper, they now have lower power than KCIT, GKCM KRR, and GKCM RF in these settings. Except for setting \ref{eq:alt1}, where GCM outperforms all other tests, the residual-based tests are outperformed by the kernel-based tests throughout. 

We conclude that even with parameter tuning of $\lambda$, distributional random forests exhibit better level and comparable or superior power than KRR in kernel-based testing. Furthermore, while the new parameter settings improved the performance of the residual-based tests under the alternative, they are still often outperformed by the kernel-based tests. 

\end{document}